\documentclass[journal]{IEEEtran}

\usepackage[numbers,sort&compress]{natbib} 
\usepackage{float}

\usepackage{multirow}
\usepackage{threeparttable}
\usepackage{booktabs}

\usepackage[ruled,linesnumbered]{algorithm2e} 
\usepackage{amsmath}
\usepackage{amsfonts}
\usepackage{amssymb}  

\usepackage{graphicx}
\graphicspath{{images/}}                     

\usepackage{hyperref}

\usepackage[utf8]{inputenc}

\usepackage{soul}
\soulregister\cite7
\soulregister\citep7
\soulregister\citet7
\soulregister\ref7
\soulregister\pageref7
\usepackage{color, xcolor} 

\begin{document}
%

\title{Semantic-aware Dense Representation Learning for Remote Sensing Image Change Detection}


\author{Hao~Chen,
        Wenyuan~Li,
        Song~Chen
        and 
        Zhenwei~Shi$^{*}$,~\IEEEmembership{Member,~IEEE}
\thanks{This work was supported by the National Natural Science Foundation of China under Grant 62125102. (Corresponding author: Zhenwei Shi)

        Hao Chen (email: justchenhao@buaa.edu.cn), 
        Wenyuan Li (email: liwenyuan@buaa.edu.cn)
        and Zhenwei Shi 
        (Corresponding Author, e-mail: shizhenwei@buaa.edu.cn) are 
        with Image Processing Center, School of Astronautics, 
        Beihang University, Beijing 100191, China,
        and with Beijing Key Laboratory of Digital Media, Beihang University,
        Beijing 100191, China, and also with State Key Laboratory of Virtual Reality
        Technology and Systems, School of Astronautics, Beihang University, Beijing
        100191, China.
        Song Chen (email: justharrychen@naver.com) is with Department of Journalism and Communications, Jeonbuk National University, Jeonju-si 54896, South Korea.
        }
}



\maketitle

\begin{abstract}
Supervised deep learning models depend on massive labeled data. Unfortunately, it is time-consuming and labor-intensive to collect and annotate bitemporal samples containing desired changes. Transfer learning from pre-trained models is effective to alleviate label insufficiency in remote sensing (RS) change detection (CD). We explore the use of semantic information during pre-training. Different from traditional supervised pre-training that learns the mapping from image to label, we incorporate semantic supervision into the self-supervised learning (SSL) framework. Typically, multiple objects of interest (e.g., buildings) are distributed in various locations in an uncurated RS image. Instead of manipulating image-level representations via global pooling, we introduce point-level supervision on per-pixel embeddings to learn spatially-sensitive features, thus benefiting downstream dense CD. To achieve this, we obtain multiple points via class-balanced sampling on the overlapped area between views using the semantic mask. We learn an embedding space where background and foreground points are pushed apart, and spatially aligned points across views are pulled together. Our intuition is the resulting semantically discriminative representations invariant to irrelevant changes (illumination and unconcerned land covers) may help change recognition. We collect large-scale image-mask pairs freely available in the RS community for pre-training. Extensive experiments on three CD datasets verify the effectiveness of our method. Ours significantly outperforms ImageNet pre-training, in-domain supervision, and several SSL methods. Empirical results indicate our pre-training improves the generalization and data efficiency of the CD model. Notably, we achieve competitive results using 20\% training data than baseline (random initialization) using 100\% data. Our code is available at \url{https://github.com/justchenhao/SaDL_CD}.

\end{abstract}


\IEEEpeerreviewmaketitle

\section{Introduction}
\label{sec:intro}

\IEEEPARstart{R}{emote} sensing (RS) image change detection (CD) is the process of identifying changes of interest in a scene based on multi-temporal RS images captured at different times \cite{Singh1989}. The definition of change varies across applications, such as urban expansion \cite{Chen2020e}, agricultural surveys \cite{Bruzzone2000}, deforestation monitoring \cite{Bem2020}, disaster damage assessment \cite{Xu2019}, etc. 

The available multi-temporal very high-resolution (VHR) RS images from satellite and aerial data provide possibilities for monitoring land cover changes at a fine scale. CD based on VHR images remains challenging due to two aspects: 1) the complexity of objects in a scene, i.e., objects of the same semantic category may exhibit varying spectral properties in different geospatial locations and seasons, 2) diverse imaging conditions, e.g., sensor characteristics and illumination variances. \cite{Bruzzone2013}. 
A good CD model can identify real changes (e.g., buildings) while ignoring irrelevant changes (e.g., illumination and unconcerned land-cover changes).

Deep learning (DL) techniques have been widely applied in the RS CD task \cite{Shi2020}. The success of DL-based supervised models heavily relies on large labeled data. Unfortunately, it is time-consuming and labor-intensive to collect and annotate bitemporal samples that contain changes of interest. 
Transfer learning is an effective solution to handle the CD labeled data insufficiency. Its basic concept is to leverage the knowledge of a pre-trained model previously trained on a large-scale dataset onto a downstream task that contains relatively small labeled data. ImageNet pre-training is commonly applied in contemporary CD models \cite{Zhang2019c, Diakogiannis2020, Zhang2020a, Chen2020, saha2020, Jiang2020a}. Considering the domain gap between natural and RS images may induce sub-optimal representations, a new trend is to pre-train on RS data to learn in-domain representations, including supervised RS pre-training \cite{Zhang2020a, Liu2020a, Wang2022b} and unsupervised RS pre-training based on self-supervised learning (SSL) \cite{Manas2021, Leenstra2021, Akiva2021, Sanchez2019}. In this work, we focus on supervised RS pre-training via leveraging existing RS data in the RS community. Different from existing supervised pre-training approaches that learn the mapping from the image to the label, we utilize the semantic supervision in a contrastive manner via incorporating it into the SSL framework.

Contrastive SSL \cite{He2019, Chen2020j} could learn meaningful representations from massive unlabeled data by pulling together representations of semantically similar samples (i.e., positive pairs) and pushing apart those of dissimilar samples (i.e., negative pairs). Very recently, contrastive methods have been introduced in the RS domain \cite{Manas2021, Leenstra2021, Akiva2021, Ayush2021, Tao2022, Kang2021, Stojnic2021, Swope2021, Jung2021, Agastya2021, Cha2021, Li2021a, Li2022a, Tarasiou2022, Li2022b} and have shown promising performance for the downstream supervised CD task \cite{Manas2021, Leenstra2021, Akiva2021}.

Most existing contrastive methods in CD express each image view into one representation vector via uniform aggregation of pixel-level embeddings. Different from the object-centered natural image (e.g., ImageNet), typically multiple small objects (e.g., buildings) are located in various positions in a real-world RS image. Contrastive learning on global representations that lack spatial information is not optimal for the downstream CD dense prediction task. Moreover, land cover spatial misalignment between different views (i.e., positive pairs) of the same image via artificial augmentations (e.g., random crop) may further hinder representation learning.

\begin{figure*}
        \centering
        \includegraphics[width=1\textwidth]{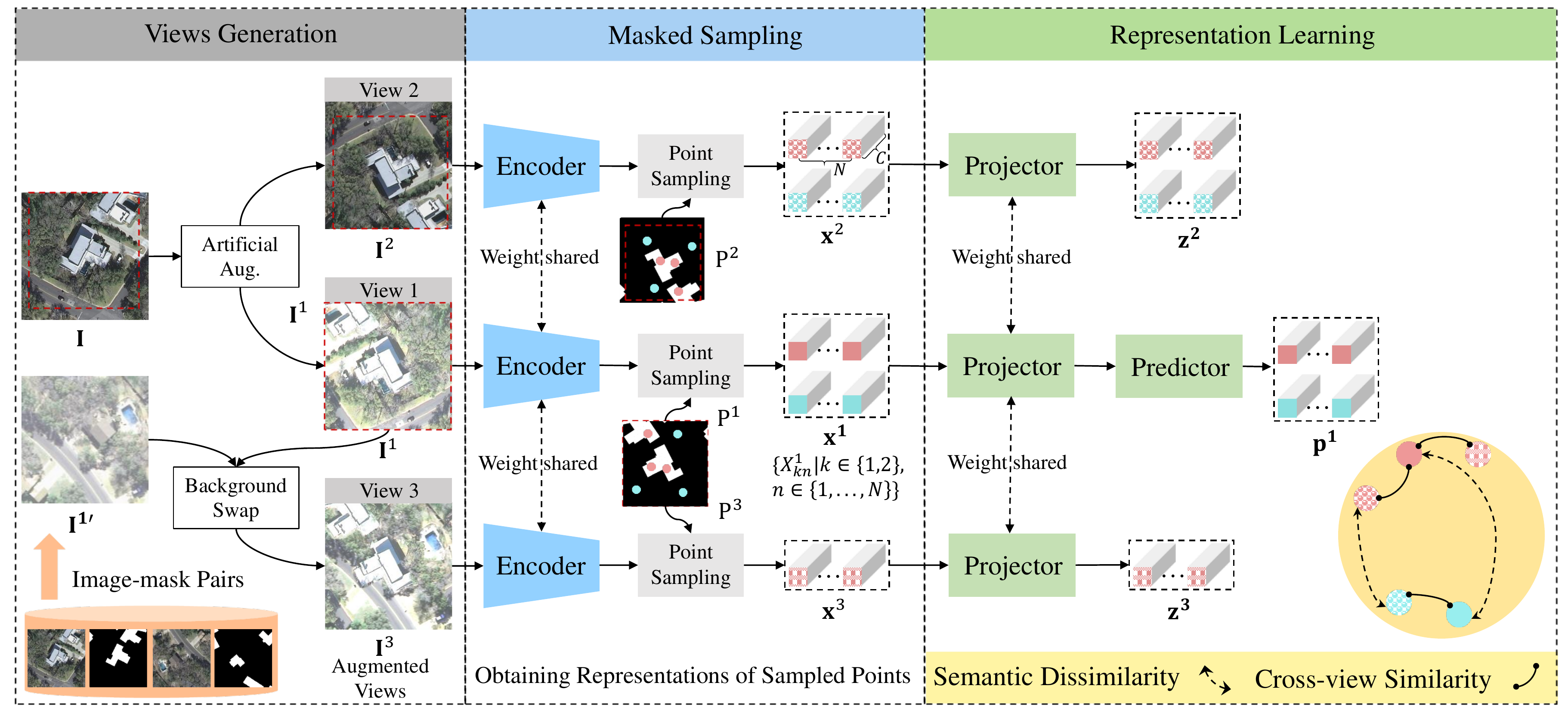}   
        \caption{Illustration of our semantic-aware dense representation learning. Our framework follows SimSiam \cite{Chen2020j} to use simple Siamese networks to learn meaningful image representations. Instead of constraining image-level consistency, we achieve point-level cross-view consistency to sustain spatially sensitive features. An equal number of points for each category are sampled from the overlapped region between views by leveraging the semantic mask. 
        We learn an embedding space where points from different semantic regions are pushed apart and those spatially aligned points across views are pulled together.
        Apart from two views via basic artificial augmentations, we generate a third view via the background swap, thus learning illumination invariant representations as well as consistent foreground representations invariant to unconcerned background changes. For simplicity, we omit the symmetrized cross-view similarity loss computed by exchanging the two views.
        }
        \label{fig:overall}
\end{figure*}

Based on the above observations, we propose semantic-aware dense representation learning for remote sensing image CD. Fig. \ref{fig:overall} illustrates the proposed method.
Instead of manipulating global representations, we constrain dense consistencies of the per-pixel image embeddings according to the location correspondence in the overlapped area between the two views. Multiple spatially aligned points across views are sampled to achieve dense correspondence. Our motivation is that the introduced point consistency may not only learn the high-level representations invariant to different augmentations (i.e., color and geometry transformations) but also sustain sensitivity to local spatial information, which helps the downstream dense prediction task.
Towards semantic-aware representation learning, we obtain class-balanced sample points from each semantic region by leveraging the semantic mask.
We follow SimSiam \cite{Chen2020j} to implement the cross-view (i.e., a positive pair) similarity of the spatially aligned points. 
We additionally employ a semantic dissimilar loss to push away foreground and background pixels in the embedding space. Such semantic discrimination design helps the downstream model to identify objects of interest (i.e. foreground).
Apart from two views by basic augmentations to the image, we generate the third view via background swap to achieve consistent foreground representations invariant to irrelevant background changes. The synthetic view of the current image is generated by blending its foreground and the background from another image using the semantic mask.

We collect a pre-training dataset that contains large-scale RS images and corresponding building semantic masks. Notably, we leverage image-mask pairs freely available in RS society \cite{Maggiori2017}. Our proposed method is evaluated on three public change detection datasets \cite{Chen2020e, Ji2019a, Peng2020}. Extensive experiments on the downstream CD task verify the effectiveness of the proposed pre-training.

The contribution of our work can be summarised as follows:
\begin{itemize}
    \item Different from traditional supervised pre-training that learns the mapping from the image to the label, we explore the use of semantic information in a representation learning framework and propose semantic-aware pre-training based on class-balanced sampling for RS image CD. Instead of manipulating image-level representations, we constrain pixel-level cross-view consistency as well as semantic discrimination to learn spatially-sensitive features, thus benefiting the downstream dense prediction CD task.
    \item Different from the self-supervised baseline, we additionally leverage semantic supervision in a contrastive manner to enhance feature discrimination ability. Apart from two views via basic augmentations, we propose a third view of a background swap version to learn representations invariant to irrelevant changes (unconcerned land covers and illumination).
    \item Extensive experiments on three CD datasets verify the effectiveness of the proposed method. Ours outperforms ImageNet pre-training, in-domain supervision, and several state-of-the-art SSL methods.

\end{itemize}

The rest of this paper is organized as follows.
Sec. \ref{sec:related-work} introduces related work of CD data-efficiency approaches and pre-training methods.
Sec. \ref{sec:method} describes our proposed semantic-aware pre-training method. 
Experimental results are given in Sec. \ref{sec:experiment}, and the Conclusion is drawn in Sec. \ref{sec:conclusion}.

\section{Related work}
\label{sec:related-work}

\subsection{Handling label insufficiency in CD} 

Training DL-based models typically rely on a massive amount of labeled data. Despite a large amount of RS data, manually collecting bitemporal images containing real changes and annotating high-quality labels could be costly. Many efforts in RS CD have been made to tackle the label insufficiency, including applying data augmentation\cite{Zhang2020b, Chen2021a, Zhang2022b,  Shi2021a, Liu2021b, Li2022d, Lei2021}, generating pseudo labels for unlabeled data via semi-supervised learning \cite{Peng2020, Hafner2022, Bandara2022}, using active learning to select a small number of informative samples \cite{Ruzicka2020, Wang2020c, Li2020d, Sahbi2021}, and transfer learning from pre-trained models \cite{Zhang2019c, Zhang2020a, Risojevi2021, Sanchez2019, Manas2021, Leenstra2021}.

Data augmentation is an effective solution to enhance the size of the training dataset. The most common way is to use transformation-based augmentations \cite{Zhang2020b, Zhang2022b,  Shi2021a, Liu2021b, Li2022d, Lei2021}, including geometric transformations (e.g., random crop, horizontal flip), color transformations, and Gaussian blur, etc. A recent advance increases the number of the positive samples (change of interest) by blending the GAN-generated instance on the appropriate spatial-temporal position of the bitemporal image \cite{Chen2021a}.

Transfer learning from a pre-trained model allows leveraging the knowledge of a source dataset to improve the data efficiency of a downstream task. ImageNet pre-training is widely used in change detection \cite{Shi2020, Zhang2019c, Diakogiannis2020, Zhang2020a, Chen2020, saha2020, Jiang2020a} and shows superior to random initialization, especially in small data regimes. Considering the domain gap between ImageNet and RS images, a new trend is to pre-train on the remote sensing data to learn the in-domain representations \cite{Zhang2020a, Liu2020a, Wang2022b, Neumann2020, Risojevi2021, Sanchez2019, Manas2021, Leenstra2021, Li2021c, Chen2022c}. Supervised RS pre-training normally learns discriminative representations from the RS image-level \cite{Zhang2020a, Wang2022b} or pixel-level \cite{Liu2020a} land-cover classification samples, while self-supervised pre-training could learn meaningful representations by making use of a vast amount of unlabeled RS data \cite{Sanchez2019, Manas2021, Leenstra2021}.

Our method falls in the supervised pre-training scope. Different from normal supervised pre-training methods that learn the mapping from image to label, we incorporate the semantic supervision in a contrastive manner in the SSL framework to enhance the feature discrimination ability. Notably, we leverage image-mask pairs freely available in RS society to guide the contrastive pre-training. 

Please note that a shorter, preliminary study of our semantic-aware representation learning has been presented in IGRASS 2022 \cite{Chen2022c}. This paper is a continuation and extension of our previous work by introducing two more pre-training designs related to the use of semantic information as well as supplementing more extensive experimental results, including more comparisons, ablations, and visualizations.

\subsection{Contrastive pre-training methods}
Self-supervised learning has attracted many researchers by virtue of its ability to learn good feature representations from massive unlabeled data to benefit various downstream tasks \cite{Liu2020c}. 
Currently, contrastive methods (e.g., SimCLR \cite{Chen2020i}, MoCo \cite{He2019, Chen2020d}, BYOL \cite{Grill2020}, SimSiam \cite{Chen2020j} and DenseCL \cite{Wang2021a}) have made significant progress in SSL. The key to contrastive approaches is to perform the instance discrimination pretext task by pulling together representations of similar images (positive pairs) and pushing apart those of dissimilar images (negative pairs). The InfoNCE loss \cite{Wu2018a} is typically applied to induce similarity between positive pairs and dissimilarity between negative pairs \cite{Chen2020i, He2019, Chen2020}. More recent attempts \cite{Grill2020, Chen2020j, Caron2021} apply simpler loss functions (e.g., negative cosine similarity and cross-entropy) that only rely on positive samples and does not use negative ones. 

Very recently, contrastive SSL methods have been introduced in RS to obtain in-domain representations. Contrastive pre-training have been shown to benefit RS downstream tasks, including land-cover classification \cite{Jean2019, Tao2020, Ayush2021, Kang2021, Stojnic2021, Swope2021, Jung2021, Agastya2021}, semantic segmentation \cite{Cha2021, Li2021a, Li2022a, Tarasiou2022}, and change detection \cite{Manas2021, Leenstra2021, Akiva2021}. Most existing methods apply InfoNCE loss \cite{Kang2021, Ayush2021, Li2022b, Swope2021, Peng2021, Li2021a, Stojnic2021, Jung2021, Agastya2021, Akiva2021} or triple loss \cite{Jean2019, Leenstra2021} on the constructed positive and negative pairs. Positive samples can be obtained by different artificial augmentations (e.g., color and geometric transformations) of the same image \cite{Stojnic2021, Agastya2021}, spatial augmentations (.i.e., geospatially overlapped images) \cite{Jean2019, Kang2021, Jung2021}, temporal augmentations (i.e., multi-temporal co-registered images) \cite{Ayush2021, Swope2021, Peng2021, Manas2021, Leenstra2021}, and modality augmentations (e.g., optical image, SAR, and semantic mask) \cite{Cha2021, Jain2022}. Negative pairs can be different samples in a mini-batch or spatially distinct images \cite{Jean2019, Peng2021}.

Contrastive methods for change detection have seldom been explored. Some attempts apply SSL directly on a small downstream change detection dataset to extract seasonal invariant features for unsupervised change detection \cite{Chen2021, Saha2021, Chen2021b}. Differently, we employ an additional relatively large-scale RS dataset to learn semantic-aware representations and then transfer knowledge from the pre-trained model to the downstream supervised CD task on a variety of CD datasets. Other more related studies that follow the normal SSL pipeline mostly evaluate the pre-trained model on the medium-resolution change detection dataset \cite{Manas2021, Leenstra2021, Akiva2021}. We instead explore pre-trained models suitable for high-resolution RS image change detection. Instead of manipulating global representations, we propose dense points consistency based on spatial correspondence between the positive pairs to better sustain the sensitivity of the spatial information, therefore benefiting the downstream dense prediction task, especially on VHR RS images that contain plenty of spatial details.
Instead of manipulating image-level representations, we constrain point-level consistencies according to spatial correspondences of the sampled pixels across positive views to better sustain the sensitivity of the spatial information, therefore benefiting the downstream dense prediction task, especially on VHR RS images that contain plenty of spatial details. Unlike the existing dense SSL method (DenseCL) that calculates similarities between representations of any two pixels to estimate pixel correspondence across views, our design utilizes the exact pixel correspondence calculated according to known geometric relations between views for learning view-invariant representations. Moreover, we leverage the semantic information in a contrastive manner to further improve the model discrimination ability to distinguish land covers of interest and other unconcerned backgrounds.
Most existing methods \cite{Chen2021, Chen2021b, Saha2021, Manas2021, Leenstra2021} treat co-registered multi-temporal images as positive samples to learn the seasonal invariant features. However, the intrinsic change of interest within the bitemporal image may induce the pre-trained model invariant to real changes, which may be unfavorable to the downstream CD task. Different from existing methods using natural bitemporal images, ours learns consistent foreground representations between one image and its synthetic version whose background is replaced by that from another image. In this way, we leverage synthetic bitemporal images to enable the model to focus more on foreground objects regardless of changes in unconcerned land covers.

\subsection{Deep learning-based models in remote sensing}
The recent few years witness the great success of deep learning techniques, e.g., convolutional neural networks (CNNs), graph convolutional networks (GCNs) \cite{Hong2021a}, and transformers \cite{Hong2022}, applied in various RS tasks \cite{Zhu2017b}, including land cover classification/semantic segmentation \cite{Hong2021a, Hong2022}, multimodal image classification \cite{Hong2021, Wu2022}, object detection \cite{Kang2022}, and change detection \cite{Shi2020, Zhang2020b, Chen2021a, Chen2020e, Zhang2022b, Shi2021a, Liu2021b, Li2022d, Lei2021, Wu2021, Qu2022, Chen2022, Bandara2022a}. CNNs have been demonstrated to be effective to extract meaningful and transferable image representations by multiple levels of abstraction, thus attracting much attention in remote sensing society. Most of the existing deep learning-based CD models \cite{Zhang2020b, Chen2021a, Chen2020e, Zhang2022b, Shi2021a, Liu2021b, Li2022d, Lei2021} are based on CNNs. GCNs originally designed for graph data and transformers for sequential data have also been employed in RS image understanding tasks \cite{Hong2021a, Hong2022}, including change detection \cite{Wu2021, Qu2022, Chen2022, Bandara2022a}, for improving model discriminative ability by leveraging their capacity of modeling long-range relations among image pixels. In this paper, instead of exploring sophisticated model structures for the CD task, we focus on the representation learning method to improve the downstream conventional CNN-based CD model.


\section{Semantic-aware Dense Representation Learning}
\label{sec:method}

In this section, we first give an overview of the proposed semantic-aware pre-training for the downstream CD task and then introduce its three main components. Finally, implementation details are given.

\subsection{Overview}

To alleviate the insufficiency of labeled CD data, we explore a supervised representation learning method by leveraging massively available data in the RS society. We collect large-scale image-mask pairs (see Sec. \ref{subsec:pre-training}) in a painless manner for contrastive pre-training. Specially, we focus on the building category as the objects of interest to perform change detection.

We propose Semantic-aware Dense representation Learning (SaDL) for remote sensing image change detection. The goal of SaDL is to learn a feature extractor that could 1) generate consistent representations invariant to illumination changes as well as changes of unconcerned land covers (i.e., background), 2) distinguish objects of interest from the background, 3) sustain spatially sensitive features to reduce the gap with the downstream dense prediction CD task.

To achieve this, we introduce the semantic mask into a self-supervised learning framework. Fig. \ref{fig:overall} illustrates the proposed SaDL.
Following SimSiam \cite{Chen2020j}, we use simple Siamese networks as the feature extractor to learn meaningful image representations without using negative samples. Our SaDL is composed of three main steps: 

1) \textbf{Views Generation}. Apart from two views of the same image via normal augmentations (i.e., color and geometry transformations), we generate a third view by swapping its background with that from another image using the semantic mask. These views can be used to learn consistent representations invariant to illumination and background changes. More generation details are given in Sec. \ref{subsec:view}.

2) \textbf{Masked Sampling}. Instead of obtaining image-level representations, our Siamese encoder converts each image view into dense per-pixel representations. We sample the same number of feature points from each semantic region in the overlapped area of the generated views. For more details see Sec. \ref{subsec:sample}.

3) \textbf{Representation Learning}. Given the sampled point representations, we employ MLP-based projection/prediction networks to perform feature transformations. We constrain point-level cross-view similarity and semantic dissimilarity to learn representations that are invariant to irrelevant changes and discriminative to foreground/background. For more details see Sec. \ref{subsec:rl}.

\subsection{Views Generation}
\label{subsec:view}

Given each input training sample (including an image $\mathbf{I}$ and a mask $\mathbf{M}$) in a mini-batch, we produce three views of the same image, including two basic transformed versions, and one background swap version.
Let $\mathcal{T}=\mathcal{T}_{g} \circ \mathcal{T}_{c}$ denotes a set of artificial augmentations, including color augmentations $\mathcal{T}_{c}$ (e.g., color jetting and Gaussian blur) and geometric augmentations $\mathcal{T}_{g}$ (e.g., random cropping and flipping). Details of our views generations are described in Algorithm \ref{alg:viewsgeneration}.

We first generate two views $\mathbf{I}^{i}$, $i\in \{1,2\}$ of $\mathbf{I}$ via different artificial augmentations:
\begin{equation}
\begin{aligned}
    \mathbf{I}^{1} &= \mathcal{T}(\mathbf{I}) = \mathcal{T}_{g}(\mathcal{T}_{c}(\mathbf{I}))\\
    \mathbf{I}^{2} &= \mathcal{T'}(\mathbf{I}) = \mathcal{T'}_{g}(\mathcal{T'}_{c}(\mathbf{I}))
\end{aligned}
\end{equation}
where $\mathcal{T}, \mathcal{T'}$ denote two random sets of artificial augmentations.

The corresponding semantic masks $\mathbf{M}^{i}$ for each view $i\in \{1,2\}$ are generated via applying the same geometric augmentations as that in the image:
\begin{equation}
\begin{aligned}
    \mathbf{M}^{1} &= \mathcal{T}_{g}(\mathbf{M}) \\
    \mathbf{M}^{2} &= \mathcal{T'}_{g}(\mathbf{M}).
\end{aligned}
\end{equation}

\textbf{View generation via background swap}. The third view of $\mathbf{I}$ is obtained by swapping the background of the first view $\mathbf{I}^{1}$ with that of another image $\mathbf{I}^{'}$. For simplicity, $\mathbf{I'}$ has the inverted index as $\mathbf{I}$ from the same mini-batch. In other words, the summation of the index of $\mathbf{I}$ and that of $\mathbf{I}^{'}$ equals the mini-batch size.
Note that we swap the common background area between $\mathbf{I}^{1'}$ and $\mathbf{I}^{1}$ to simulate the changes of unconcerned land covers.
We obtain the common background mask $\mathbf{M}_{bg}$ by calculating the intersection of the background regions in $\mathbf{M}^{1}$ and $\mathbf{M}^{1'}$.

However, directly pasting the background of $\mathbf{I}^{1'}$ on $\mathbf{I}^{1}$ using $\mathbf{M}_{bg}$ may result in undesired synthetic artifacts (see Fig. \ref{fig:composite} (d)) that could be detrimental to representation learning. Towards a realistic and effective image composite, we present a simple view synthesis method, including two main steps:

\begin{figure}
        \centering
        \includegraphics[width=0.5\textwidth]{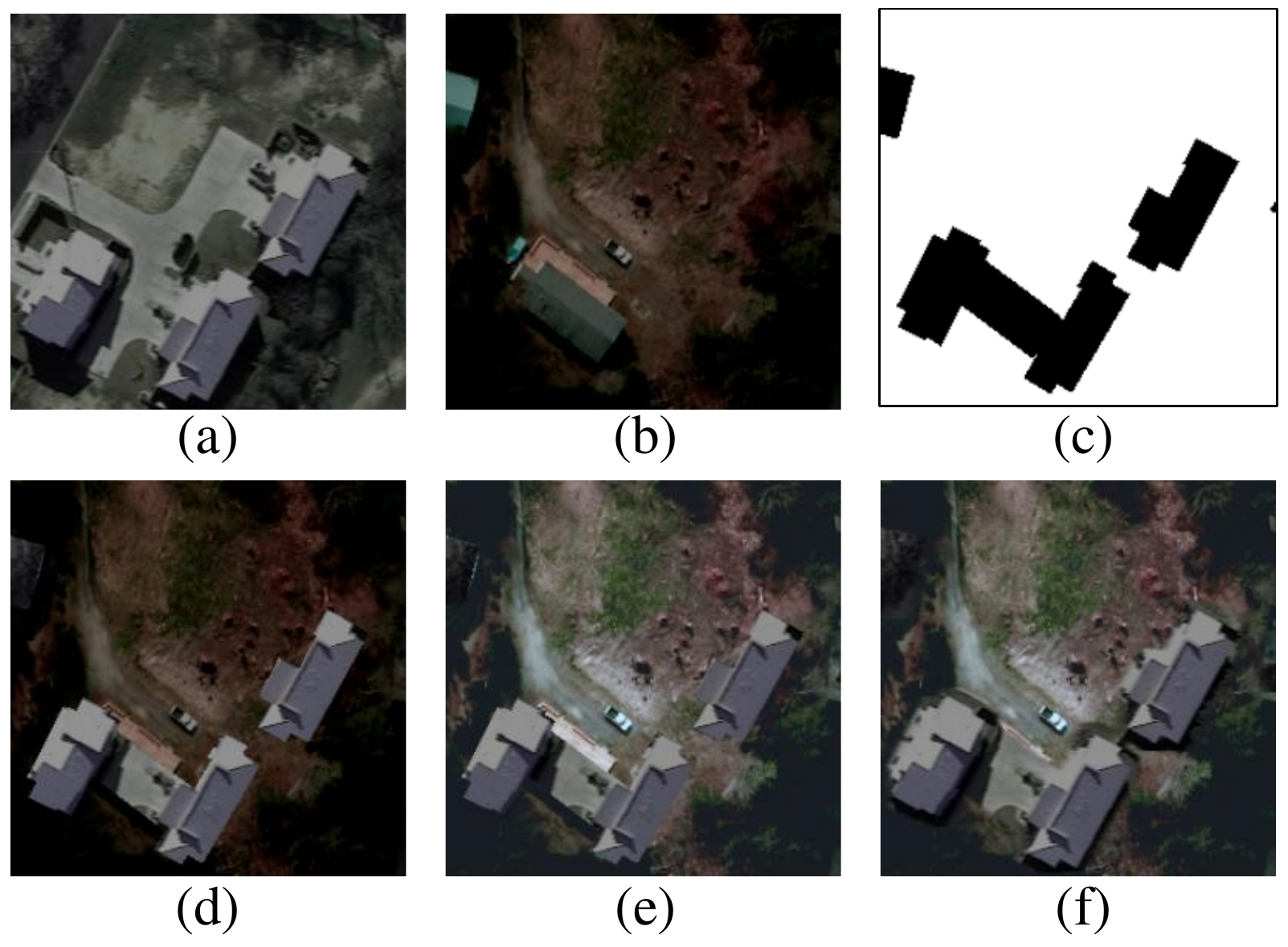}   
        \caption{Illustration of our view synthesis. View 3 is generated by swapping the background of (a) the view 1 of the current image $I^{1'}$ and that of (b) another image $I^{1'}$ using (c) their common background mask $M_{bg}$. We compare different compositions: (d) direct pasting, (e) with color transfer, (f) with color transfer + context-aware blending.}
        \label{fig:composite}
\end{figure}

1) \textbf{Color transfer}. To reduce the domain gap between $\mathbf{I}^{1'}$ (source image) and $\mathbf{I}^{1}$ (target image), we employ a simple color transfer method \cite{Huang2017a} to change the color characteristic of one image to accord with another in the image domain. The key idea is to match the mean and variance of each channel of the transformed image to that of the target image. To be formal, the transformed image is given by 
\begin{equation}
    \begin{aligned}
        \mathbf{I}^{1'}_{t}=\sigma(\mathbf{I}^{1})\left(\frac{\mathbf{I}^{1'}-\mu(\mathbf{I}^{1'})}{\sigma(\mathbf{I}^{1'})}\right)+\mu(\mathbf{I}^{1})
    \end{aligned}
\end{equation}
where $\mu(\cdot), \sigma(\cdot)$ denote the means and variances computed across spatial dimensions independently for each channel of the image.

2) \textbf{Context-aware blending}. The mere presence of the context surrounding the foreground (e.g., building) is a critical cue for object recognition. For a realistic composite, we preserve pixels near the foreground in image $\mathbf{I}^{1}$ by performing morphological erosion on the background mask. We further blur the eroded background mask using a Gaussian filter towards natural and seamless composition. The composite image $\mathbf{I}^{3}$ (view 3) is calculated by alpha blending using the eroded mask of the blurred version:
\begin{equation}
    \mathbf{I}^{3} = (1-\mathbf{M}_{bg}) \mathbf{I}^{1} + \mathbf{M}_{bg} \mathbf{I}^{1'}_{t}.
\end{equation}

Fig. \ref{fig:composite} illustrates an example of the third view synthesis. We can observe that using color transfer and context-aware blending could give a more realistic image composite.

\begin{figure}
        \centering
        \includegraphics[width=0.5\textwidth]{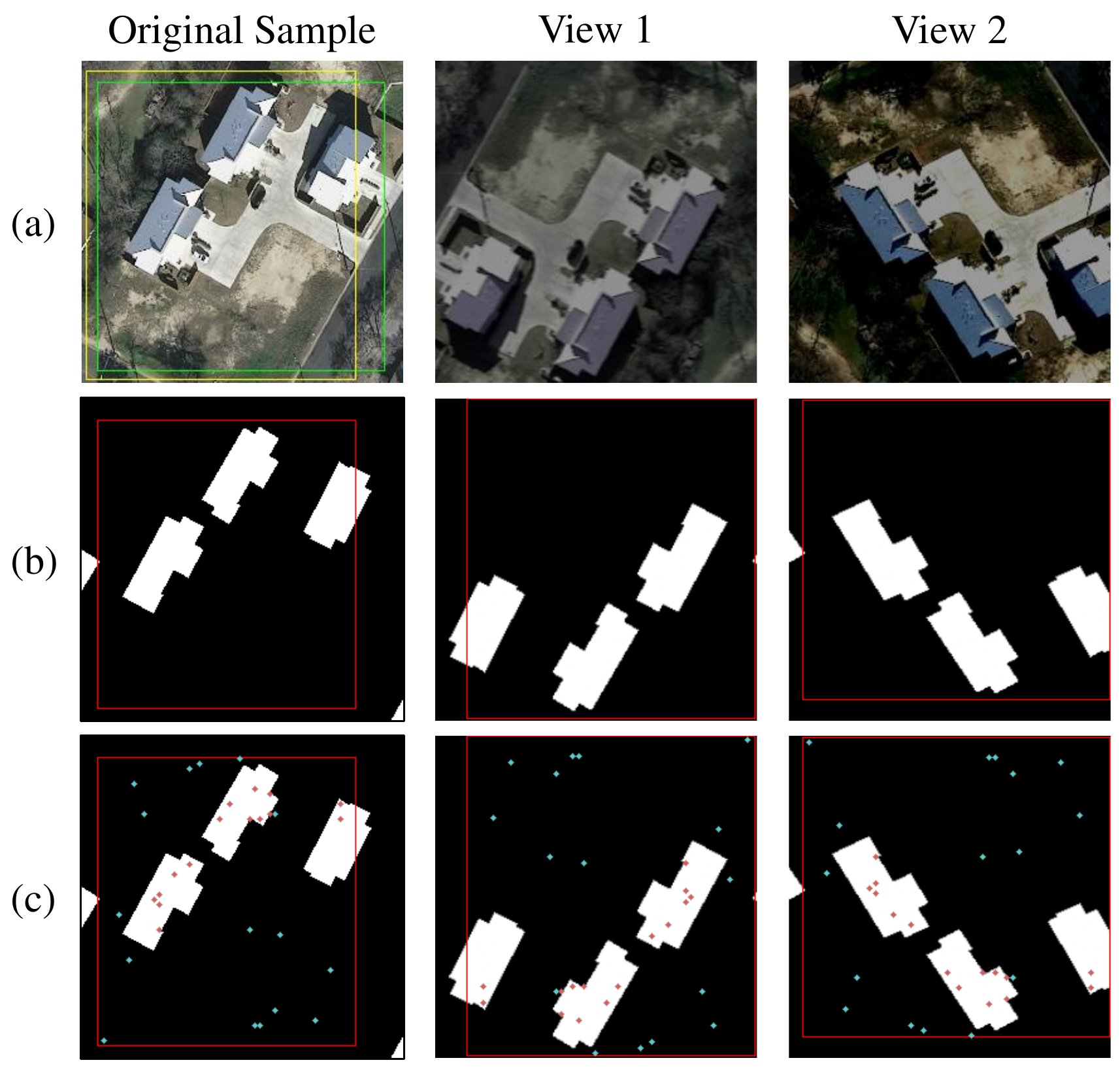}
        \caption{Illustration of our masked sampling. (a) Images. View 1/2 corresponds to the region within the green/yellow bounding box in the original image. (b) Semantic masks. The overlapped region between views is shown in red. (c) Sampled points. The same number of points for each semantic region in the overlapped area are randomly sampled. Different colors are used for each category for a better view.}
        \label{fig:views}
\end{figure}

\begin{algorithm}
\caption{Views generation for a batch of samples}
\label{alg:viewsgeneration} 

\KwIn{$\hat{\mathbf{I}}\in \mathbb{R}^{B\times H_{0}\times W_{0} \times 3}$ (a batch of input images)}
\KwIn{$\hat{\mathbf{M}}\in \mathbb{R}^{B\times H_{0}\times W_{0}}$ (a batch of corresponding semantic masks)} 
\KwOut{$\hat{\mathbf{I}}^{1},\hat{\mathbf{I}}^{2}, \hat{\mathbf{I}}^{3} \in \mathbb{R}^{B\times H_{0}\times W_{0} \times 3}$ (a batch of image view 1/2/3)}

\BlankLine

// views generation for each sample in the batch \\
\For{$b$ in $1:B$} 
{
    // generate view 1 and view 2 of the current image \\
    $\hat{\mathbf{I}}^{1}[b], \hat{\mathbf{I}}^{2}[b] \leftarrow \mathcal{T}(\hat{\mathbf{I}}[b]), \mathcal{T'}(\hat{\mathbf{I}}[b])$ \\
    $\hat{\mathbf{M}}^{1}[b], \hat{\mathbf{I}}^{2}[b] \leftarrow \mathcal{T}_{g}(\hat{\mathbf{M}}[b]), \mathcal{T'}_{g}(\hat{\mathbf{M}}[b])$ \\
}
\For{$b$ in $1:B$}
{    
    // generate view 3 of the current image \\
    // set the sample of an inverted index as the background image/mask \\
    $\mathbf{I}^{1'} \leftarrow \hat{\mathbf{I}}^{1}[B-b]$ \\ 
    $\mathbf{M}^{1'} \leftarrow \hat{\mathbf{M}}^{1}[B-b]$ \\ 
    $\mathbf{I}^{1'}_{t} \leftarrow \text{ColorTransfer}(\mathbf{I}^{1'}, \hat{\mathbf{I}}^{1}[b])$ \\ 
    // obtain the common background mask \\
    $\mathbf{M}_{bg} \leftarrow (\hat{\mathbf{M}}^{1}[b]==0) \cdot ( \mathbf{M}^{1'}==0)$ \\
    // context-aware blending \\
    $\mathbf{M}_{bg} \leftarrow \text{Erosion}(\mathbf{M}_{bg})$ \\
    $\hat{\mathbf{I}}^{3}[b] \leftarrow (1-\mathbf{M}_{bg}) \hat{\mathbf{I}}^{1} + \mathbf{M}_{bg} \mathbf{I}^{1'}_{t}$
}

\end{algorithm}

\begin{algorithm}
\caption{Dense representation learning based on masked sampling for one sample} 
\label{alg:masksample} 

\KwIn{$\mathbf{I}^{1},\mathbf{I}^{2}, \mathbf{I}^{3} \in \mathbb{R}^{B\times H_{0}\times W_{0} \times 3}$ (image view 1/2/3)}
\KwIn{$\mathbf{M} \in \mathbb{R}^{B\times H_{0}\times W_{0}}$ (semantic mask of the image)} 
\KwIn{$\mathcal{T}_{g}, \mathcal{T}_{g}^{'}$ (applied geometry augmentations for the input)} 
\KwOut{$\mathcal{L} \in \mathbb{R}$ (the loss on the input sample)}  
\BlankLine

    // reverse the area in input corresponding to view 1/2 \\
    $BB^{1},BB^{2} \leftarrow \text{Reverse}(\mathcal{T}_{g}), \text{Reverse}(\mathcal{T}_{g}^{'})$ \\
    // obtain overlapped region (BB) between views \\
    $BB \leftarrow \text{Overlap}(BB^{1}, BB^{2})$ \\
    // sampling N points in BB for each semantic category \\
    \For{$k$ in $1:2$}
    {  
        \For{$n$ in $1:N$}
        {
            sample $P_{kn} \sim \text{Uniform}(\{p=(u, v)|p \, \text{in} \, BB, \mathbf{M}[u][v]=k-1\})$
        }
    }
    // define the sampled point set \\
    $P\triangleq \{P_{kn} | n\in \{1,2,...,N\}, k\in \{1,2\}\}$ \\
    // obtain cooresponding points in each view \\
    $P^{1}, P^{2} \leftarrow \mathcal{T}_{g}(P), \mathcal{T}_{g}^{'}(P)$ \\
    $P^{3} \leftarrow \{P_{kn}^{1} | n\in \{1,2,...,N\}, k=2\}$ \\
    // coordinates downscaling \\
    $P^{1}, P^{2}, P^{3} \leftarrow P^{1}//\text{ds}, P^{2}//\text{ds}, P^{3}//\text{ds}$ \\
    //obtain point representations for each view \\
    \For{$i$ in $1:3$}
    {  
        $\mathbf{X}^{i} \leftarrow f(\mathbf{I}^{i})$ \\
        $\mathbf{x}^{i} \leftarrow \text{gather}(\mathbf{X}^{i}, P^{i})$ \\
        // obtain projections/predictions \\
        $\mathbf{z}^{i} \leftarrow g(\mathbf{x}^{i})$ \\
        $\mathbf{p}^{i} \leftarrow h(\mathbf{z}^{i})$ \\
    }

// compute losses \\
//    loss1: distinguish foreground and background \\
    \For{$i$ in $1:2$}
    {
        \For{$n$ in $1:N$}
        {
            $\mathcal{L}_{sd,n}^{i} \leftarrow \mathcal{D}(\mathbf{x}_{1n}^{i}, \mathbf{x}_{2n}^{i}) + 1$ \\
        }
    }
    $\mathcal{L}_{sd} \leftarrow \frac{1}{2N}\sum_{i=1}^{2} \sum_{n=1}^{N} \mathcal{L}_{sd,n}^{i}$ \\ 
    
//    loss2: consistency between view 1/2\\
    \For{$k$ in $1:2$}
    {
        \For{$n$ in $1:N$}
        {
            $\mathcal{L}_{s,kn}^{(1,2)} \leftarrow 1-\frac{1}{2}(\mathcal{D}(\mathbf{p}_{kn}^{1}, \text{stopgrad}(\mathbf{z}_{kn}^{2}) +$ \\ $\mathcal{D}(\mathbf{p}_{kn}^{2}, \text{stopgrad}(\mathbf{z}_{kn}^{1})))$ \\
        }
    }
    $\mathcal{L}_{s1} \leftarrow \frac{1}{2N}\sum_{k=1}^{2} \sum_{n=1}^{N} \mathcal{L}_{s,kn}^{(1,2)}$ \\ 
//    loss3: foreground consistency between view 1/3\\
        \For{$n$ in $1:N$}
        {
            $\mathcal{L}_{s,2n}^{(1,3)} \leftarrow 1-\frac{1}{2}(\mathcal{D}(\mathbf{p}_{2n}^{1}, \text{stopgrad}(\mathbf{z}_{2n}^{3}) + $\\ $\mathcal{D}(\mathbf{p}_{2n}^{3}, \text{stopgrad}(\mathbf{z}_{2n}^{1})))$ \\
        }
    $\mathcal{L}_{s2} \leftarrow \frac{1}{N} \sum_{n=1}^{N} \mathcal{L}_{s,2n}^{(1,3)}$ \\ 
    $\mathcal{L} \leftarrow \mathcal{L}_{sd} + \mathcal{L}_{s1} + \mathcal{L}_{s2}$

\end{algorithm}

\subsection{Masked Sampling}
\label{subsec:sample}

\textbf{Obtaining dense representations}.
The generated three views $\mathbf{I}^{1},\mathbf{I}^{2}, \mathbf{I}^{3}$ are then encoded by a Siamese neural network $f$ into dense representations $\mathbf{X}^{i}=f(\mathbf{I}^{i}) \in \mathbb{R}^{H \times W \times C}, i\in \{1,2,3\}$. The resulting feature map has a size of $H\times W$ and a channel dimension of $C$. Our encoder has a downsampling factor of 4. For more details see Sec. \ref{subsec:details}. 

Please note that we do not apply global average pooling to per-pixel image embeddings. Instead of manipulating image-level representations, we perform dense point-level supervision via sampling multiple points in the overlapped area across views. Our intuition is that the learned spatially-sensitive features may better reduce the gap between the pre-training and downstream dense prediction CD task that requires spatially-variant representations. Specifically, we introduce the semantic mask to achieve class-balanced sampling, i.e., an equal number of points is randomly sampled for each semantic region in the overlapped area. The details of our dense representation learning based on masked sampling on the generated views for each sample are described in Algorithm \ref{alg:masksample}.

\textbf{Overlapped region between views}. Different views (view 1 and view 2) may correspond to different spatial areas in the original image due to the difference in applied geometry augmentations. In order to perform cross-view point consistency, we first need to obtain the overlapped region between $\mathbf{I}^{1}$ and $\mathbf{I}^{2}$.
Given the known geometry augmentations ($\mathcal{T}_{g}$ and $\mathcal{T}_{g}^{'}$), we could calculate the corresponding bounding box region ($BB^{i}=(u^{i},v^{i},w^{i},h^{i})$) of view $i, i \in \{1,2\}$ in the original image $\mathbf{I}$ by utilizing the parameters of each geometry augmentation (e.g., random cropping and random flipping) in the inverse order. For example, random cropping gives the upper-left coordinate $(u,v)$, the width $w$, and the height $h$ of the cropped area in the input image. 
The overlapped region $BB$ is the intersection of the two bounding box regions $BB^{1}$ and $BB^{2}$. The region within $BB$ defines the valid sampling region and other regions outside $BB$ are invalid for the latter point sampling process.

Please note that we do not consider view 3 for calculating the overlapping region because we only take care of the foreground of view 3, which is shared with view 1.

Fig. \ref{fig:views} (a) gives an example of visualization of the view correspondence. View 1 ($\mathbf{I}^{1}$) and view 2 ($\mathbf{I}^{2}$) respectively correspond to the region within the green/yellow bounding box in the original image $\mathbf{I}$. The overlapped area between views in form of bounding boxes (red) is shown in Fig. \ref{fig:views} (b).

\textbf{Point sampling in the overlapped region}. We sample $N$ points $P_{kn}=(u_{kn}, v_{kn})$ with replacement for each semantic category $k$ in the overlapping region, where $(u_{kn}, v_{kn})$ denotes the upper-left coordinate of point $P_{kn}$ in the original image/mask ($\mathbf{I}/\mathbf{M}$). $N$ is the number of sampled points for each class. Our default $N$ is set to 16 (See Sec. \ref{subsec:details}). $k$=1 denotes the background and $k=2$ denotes the foreground class. Let $P=\{P_{kn}=(u_{kn}, v_{kn}) | n\in \{1,2,...,N\}, k\in \{1,2\}\}$ denotes the sampled point set. The corresponding points in view 1 and view 2 can be obtained by coordinate transformation under the given geometry augmentations:
\begin{equation}
\begin{aligned}
    P^{1} &= \mathcal{T}_{g}(P)\\
    P^{2} &= \mathcal{T}^{'}_{g}(P)
\end{aligned}
\end{equation}
where $P^{i}=\{P^{i}_{kn}=(u^{i}_{kn}, v^{i}_{kn}) | n\in \{1,2,...,N\}, k\in \{1,2\}\}$ denotes the set of coordinates of the sampled point in view $i, i\in \{1,2\}$. Note that we divide $P^{i}$ by ds (i.e., the downsampling factor of the encoder, defaults to 4) to match the coordinates of the per-pixel representations $X^{i}$.
The sampled points set for view 3 is given by $P^{3}=\{P^{3}_{kn}=P^{1}_{kn}) | n\in \{1,2,...,N\}, k=2\}$. 

Then for each view $i, i\in \{1,2\}$, we gather $N$ representations $\mathbf{x}^{i}_{kn}$ in the embedding space $\mathcal{V} \in \mathbb{R}^{C}$ for each semantic category $k$ by indexing points on the feature map $\mathbf{x}^{i}$ using coordinates $P^{i}$:
\begin{equation}
    \mathbf{x}^{i} = \text{gather}(\mathbf{X}^{i}, P^{i})
\end{equation}
where $\mathbf{x}^{i} = \{{x}^{i}_{kn} | n\in \{1,2,...,N\}, k\in \{1,2\}\}$ represents the set of representations of sampled points for view $i$.

Similarly, we gather $N$ representations $\mathbf{x}^{3}_{kn}$ for view 3 using sampled points $P^{3}$ that belong to the region of the foreground class (k=2). Formally, $\mathbf{x}^{3} = \{{x}^{3}_{kn} | n\in \{1,2,...,N\}, k=2\}$.

\subsection{Representation Learning}
\label{subsec:rl}

\textbf{Semantic discrimination}.
We exploit the semantic supervision in a contrastive manner via pushing apart representations of points that belong to different semantic regions in the same image.
We constrain point-level pair-wise semantic dissimilarity. Our motivation is that the learned discriminative dense features could distinguish the objects of interest and other unconcerned backgrounds, which may facilitate the downstream change recognition of the land covers of interest.

To achieve this, we minimize the cosine similarity between the background and foreground vectors (i.e., point representations from the foreground and the background, respectively) in the embedding space $\mathcal{V}$. 
For simplicity, we construct $N$ pairs of points by one-to-one correspondence between $N$ foreground points $\mathbf{x}_{2n}^{i}$ and $N$ background points $\mathbf{x}_{1n}^{i}, n\in \{1,...,N\}$ for each view $i \in \{1,2\}$. Note that we supplement 1 to cosine similarity to ensure a non-negative loss value. Our semantic dissimilar loss $\mathcal{L}_{sd,n}^{i}$ on one constructed point pair indexed $n$ for view $i$ is defined as follows: 
\begin{equation}
    \mathcal{L}_{sd,n}^{i}= \mathcal{D}(\mathbf{x}_{1n}^{i}, \mathbf{x}_{2n}^{i}) + 1
\end{equation}
where $\mathcal{D}(\cdot,\cdot)$ denotes the cosine similarity, given by:
\begin{equation}
    \mathcal{D}\left(a, b\right)=\frac{a}{\left\|a\right\|_{2}} \cdot \frac{b}{\left\|b\right\|_{2}}
\end{equation}
where $\lVert \cdot \rVert_{2}$ denotes $\ell_2$-norm, $a, b$ are two vectors of the same dimension. 

The semantic dissimilar loss for one sample is the average of $\mathcal{L}_{sd,n}^{i}$ on all the constructed point pairs in each view:
\begin{equation}
    \mathcal{L}_{sd} = \frac{1}{2N}\sum_{i=1}^{2} \sum_{n=1}^{N}\mathcal{L}_{sd,n}^{i}.
\end{equation}

\textbf{Cross-view similarity}. Given the sampled points from different views, we constrain the pair-wise similarity between point representations by utilizing the spatial correspondences of cross-view points. We aim to learn two kinds of point-level consistencies: 1) consistent representations invariant to illumination changes, and 2) consistent foreground representations regardless of background changes. We employ view 1/2 to achieve the first kind consistency and view 1/3 for the second kind.

MLP-based projection/prediction networks are applied to implement the cross-view consistency. 
For each view $i\in \{1,2,3\}$, the sampled point representations $\mathbf{x}_{kn}^{i}$ from the encoder are projected into a subspace $\mathcal{Z} \in \mathbb{R}^{C'}$ by a projection MLP head (projector) $g: \mathcal{V} \mapsto \mathcal{Z}$ to obtain projection vectors $\mathbf{z}_{kn}^{i}=g(\mathbf{x}_{kn}^{i})$. Then, a prediction MLP head (predictor) $h: \mathcal{Z} \mapsto \mathcal{Z}$ transforms the output of one view (i.e., $\mathbf{z}_{kn}^{i}$) to obtain a prediction vector (i.e., $\mathbf{p}_{kn}^{i}=h(\mathbf{z}_{kn}^{i}) \in \mathbb{R}^{C'}$) in the same space and matches it to the other view (i.e., $\mathbf{z}_{kn}^{i'}$). We minimize the negative cosine similarity of representations of corresponding points between the two views ($i/i'$). We define a symmetrized similarity loss between view $i$ and view $i'$ for $n$-$th$ point belonging to semantic category $k$ as follows:
\begin{equation}
\begin{split}
    \mathcal{L}_{s,kn}^{(i,i')} = 1-\frac{1}{2}(\mathcal{D}(\mathbf{p}_{kn}^{i}, \text{stopgrad}(\mathbf{z}_{kn}^{i'}) + \\ \mathcal{D}(\mathbf{p}_{kn}^{i'}, \text{stopgrad}(\mathbf{z}_{kn}^{i}))) 
\end{split}
\end{equation}
where stopgrad$(\cdot)$ denotes the stop-gradient operation. The encoder and projector can not receive gradient from projection vectors. Previous work \cite{Chen2020j} has empirically demonstrated the vital importance of such an operation to prevent model collapse.

The first consistency loss is the average similarity loss between view 1 and view 2 on all the sampled points $n \in \{1,...,N\}$ for each semantic category $k \in \{1,2\}$:
\begin{equation}
    \mathcal{L}_{s1} =\frac{1}{2N}\sum_{k=1}^{2}\sum_{n=1}^{N}\mathcal{L}_{s,kn}^{(1,2)}.
\end{equation}

Similarly, the second consistency loss is the average similarity loss between view 1 and view 3 on all the sampled points $n \in \{1,...,N\}$ that belong to the foreground ($k$=2):
\begin{equation}
    \mathcal{L}_{s2} =\frac{1}{N}\sum_{n=1}^{N}\mathcal{L}_{s,2n}^{(1,3)}.
\end{equation}

\textbf{Overall loss function}. We calculate the summation of the semantic dissimilar loss and the two kinds of similarity losses to obtain our overall loss for each sample:
\begin{equation}
    \mathcal{L} = \mathcal{L}_{sd}+\mathcal{L}_{s1}+\mathcal{L}_{s2}.
\end{equation}
Note that the final loss is averaged over all samples in a mini-batch.

\subsection{Implementation Details}
\label{subsec:details}

\textbf{Encoder.} 
We use the ResNet-18 \cite{He2016} backbone (without fully connected layers and global pooling) equipped with Feature Pyramid Networks (FPN) \cite{Lin2017a} as our default encoder $f$ to generate high-resolution and high-level per-pixel features for each image view. 
The ResNet-18 has 5 stages, each with downsampling by 2. Through the feedforward computation of the backbone, we obtain a feature hierarchy consisting of feature maps at different scales from the activation outputs of the last four stages. The FPN takes as input the feature hierarchy and progressively upsamples the coarse-resolution feature maps while merging them with the corresponding intermediate feature maps by element-wise addition. Each intermediate feature map undergoes a $1\times1$ convolutional layer to transform to the same channel dimension $C$ before addition. The resulting feature map has a 1/4 spatial size of the input image. The channel dimension $C$ of output features is set to 256.

\textbf{Projector.} The projector $g$ is a 2-layer MLP with a hidden dimension of 2048 and an output dimension of 1024. BN and ReLU are added between fully connected (fc) layers.

\textbf{Predictor.} The predictor $h$ is a 2-layer MLP with a hidden dimension of 256 and an output dimension of 1024. BN and ReLU are also added between the two fc layers.

\textbf{Data augmentation.} We set probability of color jittering to 0.8, with \{brightness, contrast, saturation, hue\} strength of \{0.4, 0.4, 0.4, 0.1\}, and the probability of Gaussian blurring to 0.5 with a kernel standard deviation in [0.1, 2.0]. Geometric augmentations include random vertical flip, random horizontal flip, and random resized crop with a scale in [0.8, 1.0]. Note that we do not follow the common scale \cite{Chen2020j} for the random crop (i,e., [0.2, 1.0]) because the two cropped regions with a small scale may not contain an overlapped area, which may be detrimental to the effective sampling of spatially-aligned points across views.

\textbf{Optimizer}. We use Stochastic Gradient Descent (SGD) with a momentum hyperparameter of 0.9 and a weight decay of 0.0005. The initial learning rate is 0.01 and decays to zero at the end of training. We follow the "poly" learning rate schedule \cite{Chen2018} with a decay coefficient of 0.9. The default number of pre-training epochs $N_{e}$ is 200 (see Sec. \ref{subsec:parameter}). Due to capacity limitations of memory, the default batch size is set to 64. Our models are implemented on PyTorch and trained using a single NVIDIA RTX 3090 GPU. 

\section{Experimental Results}
\label{sec:experiment}

\subsection{Pre-training Setup}
\label{subsec:pre-training}

\textbf{Pre-training Dataset}. We leverage image-mask pairs from the existing Inria building segmentation dataset \cite{Maggiori2017}, which contains 180 labeled RS images, each size of $5000\times 5000$ and spatial resolution of 0.3 m. It provides pixel-level annotations, including the building and non-building categories.
We cut the original samples into small patches of size $256\times 256$ with no overlap and remove those without building regions. The resulting dataset contains more than 45k patch samples. We randomly split it into training (80\%) and validation (20\%) sets. We additionally obtain the co-registered image patch of the corresponding geospatial region via Google Earth as the temporal augmentation ($t_{2}$) for each image ($t_{1}$) in the dataset. 
For a fair comparison, the natural augmentations (bitemporal images) are used to train some other related pre-training methods that learn seasonal invariant features. Please note that the semantic mask may not perfectly match the image of $t_{2}$ due to land-cover changes (e.g., building construction and demolition) over time, we do not include the augmented temporal image in our pre-training framework.

\subsection{Downstream Change Detection Setup}
\subsubsection{Change detection datasets}
To evaluate the proposed pre-trained model, we conduct experiments on the following three change detection datasets:

\textbf{LEVIR-CD} \cite{Chen2020e}: a widely-used building change detection dataset. LEVIR-CD contains 637 pairs of bitemporal VHR RS images, each with a size of $1024\times 1024$ and a spatial resolution of 0.5m. We follow the default train-valid-test split \cite{Chen2020e}. Each image is cropped with no overlap into small patches of size $256\times 256$. Finally, we obtain 7120/1024/2048 samples for training/validation/testing, respectively.

\textbf{WHU-CD} \cite{Ji2019a}: a public building change detection dataset that includes one pair of optical VHR RS aerial images with a size of $32507\times15354$  and 0.075m spatial resolution. Similarly, we crop the original sample into small patches of size $256\times 256$ with no overlap. We randomly split the dataset into training, validation, and testing sets, with 6096/762/762 samples, respectively.

\textbf{Guangzhou-CD} \cite{Peng2020}: a building change detection dataset that covers several suburb areas of Guangzhou City, China. It contains 19 VHR RGB image pairs (0.55m) with sizes ranging from $1006 \times 1168$ to $4936 \times 5224$ pixels. Similarly, the image pairs are cropped into $256\times256$ non-overlapping patches. As the dataset provider does not give a default dataset split, we randomly split the dataset into three parts, with 2882/361/360 samples for training, validation, and testing, respectively. 

\subsubsection{Change detection networks}
We employ a simple yet effective CD model \cite{Chen2021a} which consists of a Siamese feature extractor (deep FCN) to extract high-resolution semantic feature maps for bitemporal patches, a distance metric to calculate the Feature Difference Images (FDI) between the two patches, and a relatively shallow FCN to give the change probability maps. Differently, we employ a light FPN-based decoder head \cite{Lin2017a} for the feature extractor with the encoder-decoder architecture. Our change detection networks employ a ResNet-18 backbone. We use the cross-entropy loss to optimize our CD model. Please refer to \cite{Chen2021a} for more details.

\textbf{Fine-tuning details}. 
We fine-tune the downstream CD networks using the pre-trained models. For a fair comparison, we only initialize parameters of the backbone (i.e., ResNet-18) of the CD networks by those from the pre-trained model. Note that parameters of other components in the CD networks are initialized with zero-mean normal distribution with a standard deviation of 0.02.

\textbf{Training details}. We apply data augmentation, including random flip, and Gaussian blur. Batch size is set to 8. SGD is used for model optimization. We set the momentum to 0.9 and the weight decay to 0.0005. The learning rate starts at 0.01 and linearly decays to 0 until 200 epochs. We perform validation after each training epoch, and use the best model on the validation set for evaluation on the test set.

\textbf{Evaluation Metrics}. The F1-score regarding the change category is used as the evaluation indices. Additionally, precision, recall, and Intersection over Union (IoU) of the change category are reported. These metrics are defined as follows: 
\begin{equation}
\begin{split}
        & \text{F1} = \frac{2}{\text{recall}^{-1}+\text {precision}^{-1}} \\
        & \text{precision} = \frac{\text{TP}}{\text{TP} + \text{FP}} \\
        & \text{recall} = \frac{\text{TP}}{\text{TP} + \text{FN}} \\
        & \text{IoU} = \frac{\text{TP}}{\text{TP} + \text{FN} + \text{FP}} \\
\end{split}
\end{equation}
where $\text{TP, FP, FN}$ denote the number of true positive, false positive, and false negative, respectively.

\subsection{Overall Comparison}
\label{ssec:comparison}
To evaluate the effectiveness of the proposed pre-training method, we make a comparison with the random initialization, ImageNet pre-training, in-domain supervised pre-training, and several self-supervised pre-training methods:
\begin{itemize}
    \item \textit{Random initialization}. All the parameters of the CD networks are initialized by the zero-mean normal distribution with a deviation of 0.02.
    \item \textit{ImageNet pre-training}. The backbone (ResNet-18) of the CD model is initialized with the ImageNet pre-training.
    \item \textit{In-domain supervised pre-training (In-domain Sup.)}: We employ an FCN-based semantic segmentation network with a ResNet-18 backbone and an FPN head \cite{Lin2017a}, supervised by the image-mask pairs of our pre-training dataset. We follow the same optimization configuration as that for CD networks. We perform the model evaluation in the validation set after each training epoch and save the best model (backbone) for the downstream task.
    \item \textit{SimSiam} \cite{Chen2020j}: A popular SSL method without using negative samples. SimSiam employs the stop-gradient and predictor to prevent model collapse.
    \item \textit{MoCo-v2} \cite{Chen2020d}: A competitive SSL model that formulates contrastive learning as a dictionary look-up problem where a query should be similar to its matching key and dissimilar to others in a dynamic queue covering a large set of negative samples.
    \item \textit{DenseCL} \cite{Wang2021a}: A dense contrastive learning method that constrains pixel-level cross-view consistencies where the pixel correspondence is computed via Ssing pair-wise similarities between any two pixels from different views.
    \item \textit{CMC} \cite{Cha2021}: A recent SSL method in RS that takes the semantic mask as an additional view along with the optical/SAR image to learn consistent cross-modal representations under the framework of contrastive multiview coding \cite{Tian2020a}. Note that we implement multi-modal representation learning with image-mask pairs of our pre-training dataset. 
    \item \textit{SeCo} \cite{Manas2021}: A very recent SSL method that is based on MoCo-v2 and further exploits the use of multi-temporal RS images as natural augmentations to learn seasonal invariant/variant representations. We implement it using the bitemporal images in our pre-training dataset.
    
\end{itemize}

For a fair comparison, we use our pre-training dataset to train these supervised/self-supervised methods, i.e., in-domain supervision, SimSiam, MoCo-v2, DenseCL, CMC, and SeCo. The SSL methods are implemented using their public codes with default hyperparameters. Note that all the compared models are initialized with ImageNet pre-trained weights as our SaDL does.

Each pre-trained model is evaluated by the performance of the downstream CD task. Tab. \ref{tab:comparison_all} reports the overall comparisons of our proposed method and other pre-trained models on the LEVIR-CD, WHU-CD, and Guangzhou-CD datasets, respectively. Precision, recall, F1-score, and IoU of the CD networks on the three CD test sets are given. To evaluate the data-efficient performance, we set several data regimes: 1\%, 5\%, 20\%, and 100\%, where each number represents the proportion of available labeled data in the original training set.

From Tab. \ref{tab:comparison_all}, we can observe that the proposed method consistently outperforms the compared pre-trained models in terms of the F1-score across different data regimes on the three datasets. Specifically, our SaDL pre-training is significantly superior to random initiation. It is notable that we can achieve comparable or even better results with only 20\% training data than the random initiation baseline using 100\% training data. It indicates that our pre-training can effectively alleviate the labeled data insufficiency problem.

ImageNet pre-training serves as a strong baseline due to its powerful general representations learned from the large-scale labeled ImageNet dataset. The empirical results demonstrate that our method overachieves ImageNet pre-training across different data conditions on the downstream CD task. For instance, Tab. \ref{tab:comparison_all} (a) shows that compared to ImageNet pre-training, we achieve considerable improvements of more than 20\% of the F1-score in the LEVIR-CD test set under small data regimes (1\% and 5\%).

\textbf{Comparison to in-domain supervision}.
Typically, supervised RS pre-training obtains in-domain discriminative representations by learning the mapping from RS images to labels. Tab. \ref{tab:comparison_all} shows the in-domain supervision is consistently inferior to our SaDL under each data configuration of the three datasets, even not comparable to ImageNet pre-training in some data regimes. Note that we use the same amount of pre-training data for SaDL and in-domain supervision. The quantitative results indicate that our SaDL can learn more effective transferable features than the traditional approach. We can observe that in-domain supervision shows relatively poorer results on the WHU-CD/Guangzhou-CD datasets than on the LEVIR-CD dataset. It may be due to the subdomain diversity, i.e., different RS datasets contain land-covers of diverse appearances and different spatial resolutions. The traditional supervised pre-training may overfit patterns of a certain subdomain and lack transferability to another.

\textbf{Comparison to other SSL pre-training}.
Tab. \ref{tab:comparison_all} also shows that our method consistently outperforms several SSL methods, including two popular image-level representation learning methods, one dense contrastive learning method, and two recent SSL methods in the RS domain. We can observe that the normal dense contrastive learning (DenseCL) can even be inferior to the image-level counterpart (MoCo) in many data conditions. It may due to the discrepancy between the commonly used ImageNet image and the uncurated RS image. Unlike the ImageNet data that mainly contains one single object in the center of the image, multiple similar foreground objects may exist at different spatial locations in an RS image. The intrinsic characteristic of RS data brings difficulty in computing pixel correspondence across views in DenseCL. The resulting incorrect correspondence may hinder learning meaningful representations. On the contrary, we utilize the exact pixel correspondence for representation learning by leveraging spatial correspondences of the pixels in the overlapped area between views.

The comparison with CMC indicates that we exploit a better approach to using the semantic mask as additional information in a self-supervised learning framework to improve feature representations. We argue that directly using the mask as an augmented view along with the image views may not be an optimal solution.
In \cite{Cha2021}, the mask is encoded into one vector that losses spatial details. Semantic spatial relations have not been fully utilized in such a scenario. Instead, we leverage semantic masks to push away representations of pixels from different semantic regions to learn semantic-aware features that can distinguish the foreground or background pixels, thus benefiting the recognition of objects of interest in the downstream CD task.

\begin{figure*}
        \centering
        \includegraphics[width=\textwidth]{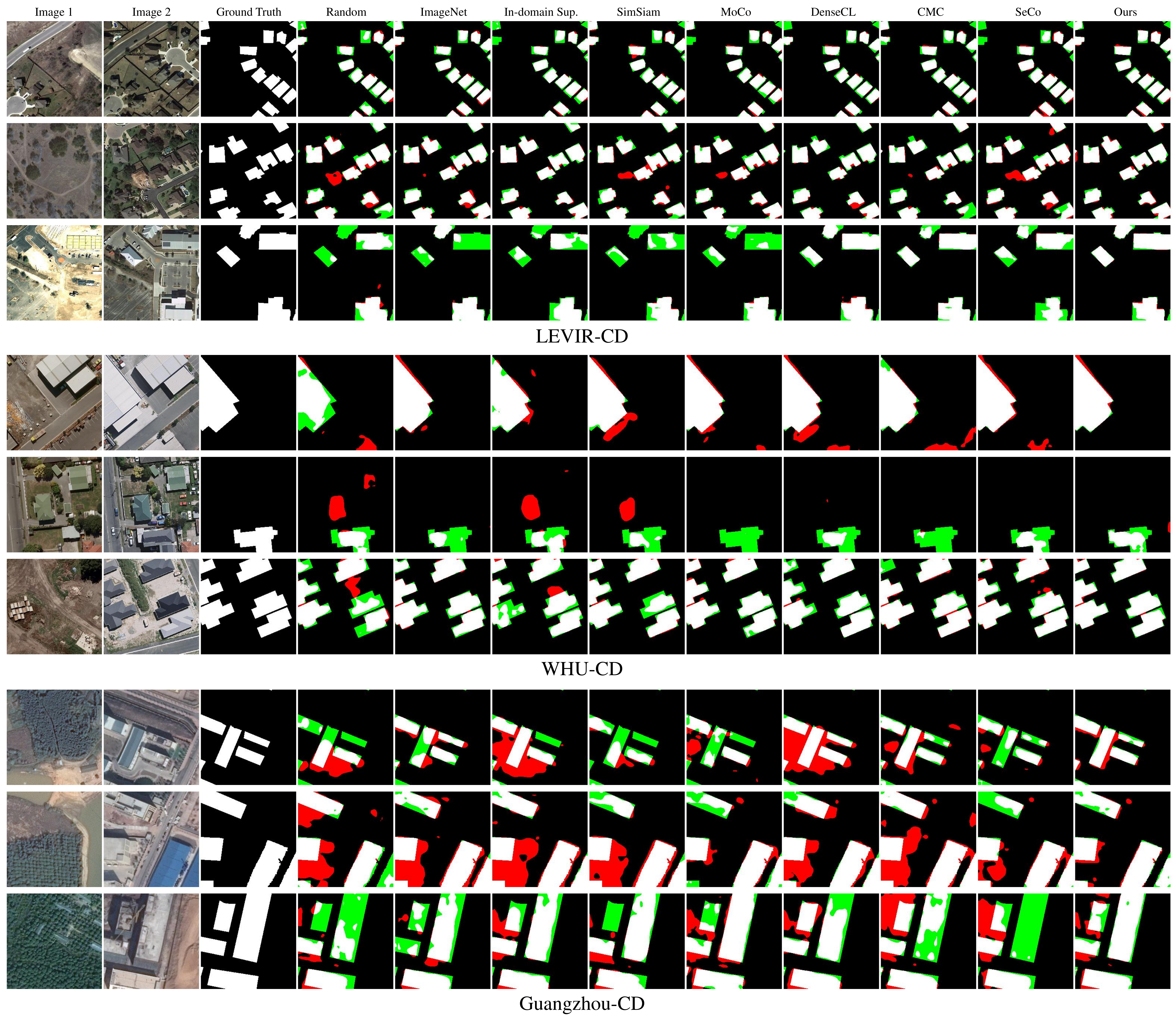}   
        \caption{Visualization comparison of the CD model fine-tuned from different pre-trained models on the LEVIR-CD/WHU-CD/Guangzhou-CD test sets. The CD model is trained under a 20\% data regime. For a better view, white for true positives, black for true negatives, red for false positives, and green for false negatives.}
        \label{fig:predictions}
\end{figure*}

Visualization comparisons of the CD predictions on the three datasets for different pre-trained methods are also given in Fig. \ref{fig:predictions}. We use different colors for a better view, where white for true positives, black for true negatives, red for false positives, and green for false negatives. We could observe that the CD model using our pre-trained model achieves more accurate predictions than those of its counterparts.

\begin{table*}
    \centering
    \caption{Overall comparisons of different pre-training methods on the three CD dataset. Precision/recall/F1/IoU of the CD model on the LEVIR-CD/WHU-CD/Guangzhou-CD test set are reported. The highest classification accuracy in each data regime is marked in bold.}
    \begin{minipage}[t]{\linewidth}
    \centering
    \centerline{\normalsize{(a) LEVIR-CD}}
    \resizebox{1\textwidth}{!}{
        \begin{tabular}{c|c|c|c|c}
        \toprule
        \multicolumn{1}{c}{} &
        \multicolumn{1}{|c|}{1\%} & \multicolumn{1}{|c|}{5\%}  & \multicolumn{1}{|c|}{20\%}  &  \multicolumn{1}{c}{100\%} \\
        & Precision / Recall / F1 / IoU &
        Precision / Recall / F1 / IoU &
        Precision / Recall / F1 / IoU &
        Precision / Recall / F1 / IoU \\
        \midrule
        Random &
        49.83 / 13.56 / 21.32 / 11.93 &	
        73.54 / 34.58 / 47.05 / 30.76 &
        83.00 / 77.76 / 80.30 / 67.08 &
        90.50 / 84.94 / 87.63 / 77.98 \\
        ImageNet & 
        30.80 / 20.11 / 24.33 / 13.85 &
        81.40 / 54.35 / 65.18 / 48.35 &
        89.59 / 82.26 / 85.76 / 75.08 &
        91.29 / 85.25 / 88.16 / 78.83 \\
        In-domain Sup. &  
        79.43 / 17.96 / 29.29 / 17.16 &
        86.25 / 61.73 / 71.96 / 56.20 &
        \textbf{90.91} / 81.02 / 85.68 / 74.95 &
        91.22 / 85.11 / 88.06 / 78.67\\
        SimSiam \cite{Chen2020j} & 
        71.90 / 13.29 / 22.44 / 12.64 &
        82.46 / 44.41 / 57.73 / 40.58 &
        88.97 / 76.46 / 82.24 / 69.84 &
        91.47 / 84.78 / 88.00 / 78.57 \\
        MoCo-v2 \cite{Chen2020d} &
        58.15 / 20.30 / 30.09 / 17.71 &
        74.78 / 71.20 / 72.95 / 57.42 &
        88.78 / 80.45 / 84.41 / 73.03 &
        91.59 / 85.06 / 88.21 / 78.90\\
        DenseCL \cite{Wang2021a} &
        72.45 / 28.76 / 41.17 / 25.92 &
        \textbf{91.51} / 58.62 / 71.46 / 55.60 &
        89.33 / 82.09 / 85.56 / 74.76 &
        \textbf{94.63} / 79.98 / 86.69 / 76.51\\
        CMC \cite{Cha2021} &
        79.43 / 10.42 / 18.42 / 10.15 &
        84.45 / 56.40 / 67.63 / 51.09 &
        90.30 / 77.94 / 83.67 / 71.92 &
        90.73 / 86.59 / 88.61 / 79.55\\
        SeCo \cite{Manas2021} &
        68.63 / 29.01 / 40.78 / 25.62 &
        84.74 / 61.43 / 71.23 / 55.31 &
        84.34 / 83.70 / 84.02 / 72.44 &
        91.03 / 85.68 / 88.27 / 79.01\\
        \midrule
        Ours &
        \textbf{82.38} / \textbf{42.52} / \textbf{56.09} / \textbf{38.98} &
        88.35 / \textbf{72.16} / \textbf{79.44} / \textbf{65.90} &
        89.58 / \textbf{85.05} / \textbf{87.25} / \textbf{77.39} &
        90.91 / \textbf{86.66} / \textbf{88.74} / \textbf{79.75} \\
        \bottomrule
       \end{tabular}
    }
    \end{minipage}
    \vspace{2mm}
    
    \begin{minipage}[t]{\linewidth}
    \centering
    \centerline{\normalsize{(b) WHU-CD}}
    \resizebox{1\textwidth}{!}{
        \begin{tabular}{c|c|c|c|c}
        \toprule
        \multicolumn{1}{c}{} &
        \multicolumn{1}{|c|}{1\%} & \multicolumn{1}{|c|}{5\%}  & \multicolumn{1}{|c|}{20\%}  &  \multicolumn{1}{c}{100\%} \\
        & Precision / Recall / F1 / IoU &
        Precision / Recall / F1 / IoU &
        Precision / Recall / F1 / IoU &
        Precision / Recall / F1 / IoU \\
        \midrule
        Random &
        37.03 / 41.76 / 39.25 / 24.42 &	
        67.99 / 64.40 / 66.14 / 49.41 &
        70.07 / 72.68 / 71.35 / 55.46 &
        85.82 / 83.20 / 84.49 / 73.15 \\
        ImageNet & 
        31.37 / 33.55 / 32.42 / 19.35 &
        80.12 / 67.92 / 73.52 / 58.12 &
        87.82 / 76.87 / 81.98 / 69.46 &
        91.11 / 86.14 / 88.55 / 79.46 \\
        In-domain Sup. &  
        43.41 / 60.49 / 50.55 / 33.82 &
        74.50 / 66.06 / 70.03 / 53.88 &
        70.79 / 72.34 / 71.56 / 55.71 &
        84.10 / 82.42 / 83.25 / 71.31\\
        SimSiam \cite{Chen2020j} & 
        37.86 / 47.38 / 42.09 / 26.65 &
        60.37 / 73.03 / 66.10 / 49.36 & 
        70.89 / 76.11 / 73.41 / 57.99 & 
        86.30 / 83.14 / 84.69 / 73.45 \\
        MoCo-v2 \cite{Chen2020d} &
        71.36 / 57.44 / 63.65 / 46.68 &
        76.45 / 67.21 / 71.53 / 55.68 & 
        85.30 / 76.36 / 80.58 / 67.48 & 
        91.36 / 84.69 / 87.90 / 78.41\\
        DenseCL \cite{Wang2021a} &
        41.02 / 46.45 / 43.56 / 27.85 &
        68.78 / 69.05 / 68.92 / 52.58 &
        84.14 / 74.37 / 78.95 / 65.22 &
        89.38 / 84.64 / 86.94 / 76.90\\
        CMC \cite{Cha2021} &
        50.89 / 45.32 / 47.94 / 31.53 &
        73.69 / 67.54 / 70.49 / 54.42 &
        86.28 / 74.54 / 79.98 / 66.64 &
        89.96 / 85.47 / 87.66 / 78.03\\
        SeCo \cite{Manas2021} &
        67.39 / 41.92 / 51.68 / 34.85 &
        78.81 / 66.52 / 72.15 / 56.43 &
        83.93 / 76.43 / 80.00 / 66.67 &
        89.29 / 86.93 / 88.09 / 78.72\\
        \midrule
        Ours &
        \textbf{78.50} / \textbf{66.29} / \textbf{71.88} / \textbf{56.10} &
        \textbf{86.45} / \textbf{75.04} / \textbf{80.34} / \textbf{67.15} &
        \textbf{90.15} / \textbf{81.37} / \textbf{85.53} / \textbf{74.73} &
        \textbf{93.10} / \textbf{87.05} / \textbf{89.98} / \textbf{81.78} \\
        \bottomrule
       \end{tabular}
    }
    \end{minipage}
    \vspace{2mm}
    
    \begin{minipage}[t]{\linewidth}
    \centering
    \centerline{\normalsize{(c) Guangzhou-CD}}
    \resizebox{1\textwidth}{!}{
        \begin{tabular}{c|c|c|c|c}
        \toprule
        \multicolumn{1}{c}{} &
        \multicolumn{1}{|c|}{1\%} & \multicolumn{1}{|c|}{5\%}  & \multicolumn{1}{|c|}{20\%}  &  \multicolumn{1}{c}{100\%} \\
        & Precision / Recall / F1 / IoU &
        Precision / Recall / F1 / IoU &
        Precision / Recall / F1 / IoU &
        Precision / Recall / F1 / IoU \\
        \midrule
        Random &
        40.42 / 51.63 / 45.34 / 29.32 &
        62.33 / 50.03 / 55.51 / 38.42 &
        72.89 / 54.51 / 62.37 / 45.32 &
        91.80 / 65.02 / 76.12 / 61.45 \\
        ImageNet & 
        52.68 / 38.36 / 44.39 / 28.53 &
        62.54 / \textbf{57.82} / 60.09 / 42.94 &
        88.93 / 59.59 / 71.36 / 55.48 &
        94.99 / 76.81 / 84.94 / 73.82\\
        In-domain Sup. &  
        36.54 / \textbf{51.99} / 42.92 / 27.32 &
        66.53 / 42.04 / 51.52 / 34.70 &
        69.82 / 60.43 / 64.79 / 47.91 &
        88.96 / 73.11 / 80.26 / 67.03\\
        SimSiam \cite{Chen2020j} & 
        42.40 / 51.09 / 46.34 / 30.16 &
        73.02 / 43.49 / 54.51 / 37.47 &
        67.29 / 61.14 / 64.07 / 47.13 &
        91.80 / 72.60 / 81.08 / 68.18 \\
        MoCo-v2 \cite{Chen2020d} &
        50.73 / 48.56 / 49.62 / 33.00 &
        67.33 / 56.56 / 61.48 / 44.38 &
        81.24 / 61.80 / 70.20 / 54.08 &
        90.23 / 76.69 / 82.91 / 70.81\\
        DenseCL \cite{Wang2021a} &
        37.06 / 51.16 / 42.98 / 27.37 &
        58.69 / 40.78 / 48.12 / 31.69 &
        77.36 / 57.88 / 66.21 / 49.49 &
        93.19 / 71.36 / 80.83 / 67.82\\
        CMC \cite{Cha2021} &
        46.77 / 38.94 / 42.50 / 26.98 &
        75.17 / 53.37 / 62.42 / 45.37 &
        73.97 / 66.46 / 70.01 / 53.86 &
        92.20 / 76.72 / 83.75 / 72.04\\
        SeCo \cite{Manas2021} &
        46.75 / 50.62 / 48.61 / 32.11 &
        63.50 / 56.70 / 59.91 / 42.76 &
        80.08 / 60.62 / 69.00 / 52.68 &
        \textbf{92.51} / 76.98 / 84.03 / 72.46\\
        \midrule
        Ours &
        \textbf{57.96} / 47.79 / \textbf{52.39} / \textbf{35.49} &
        \textbf{80.04} / 56.00 / \textbf{65.89} / \textbf{49.13} &
        \textbf{85.32} / \textbf{67.06} / \textbf{75.09} / \textbf{60.12} &
        92.13 / \textbf{81.06} / \textbf{86.24} / \textbf{75.81} \\
        \bottomrule
       \end{tabular}
    }
    \end{minipage}
    
\label{tab:comparison_all}
\end{table*}

\begin{figure*}
\begin{minipage}[t]{0.33\linewidth}
\centering
\includegraphics[width=\textwidth]{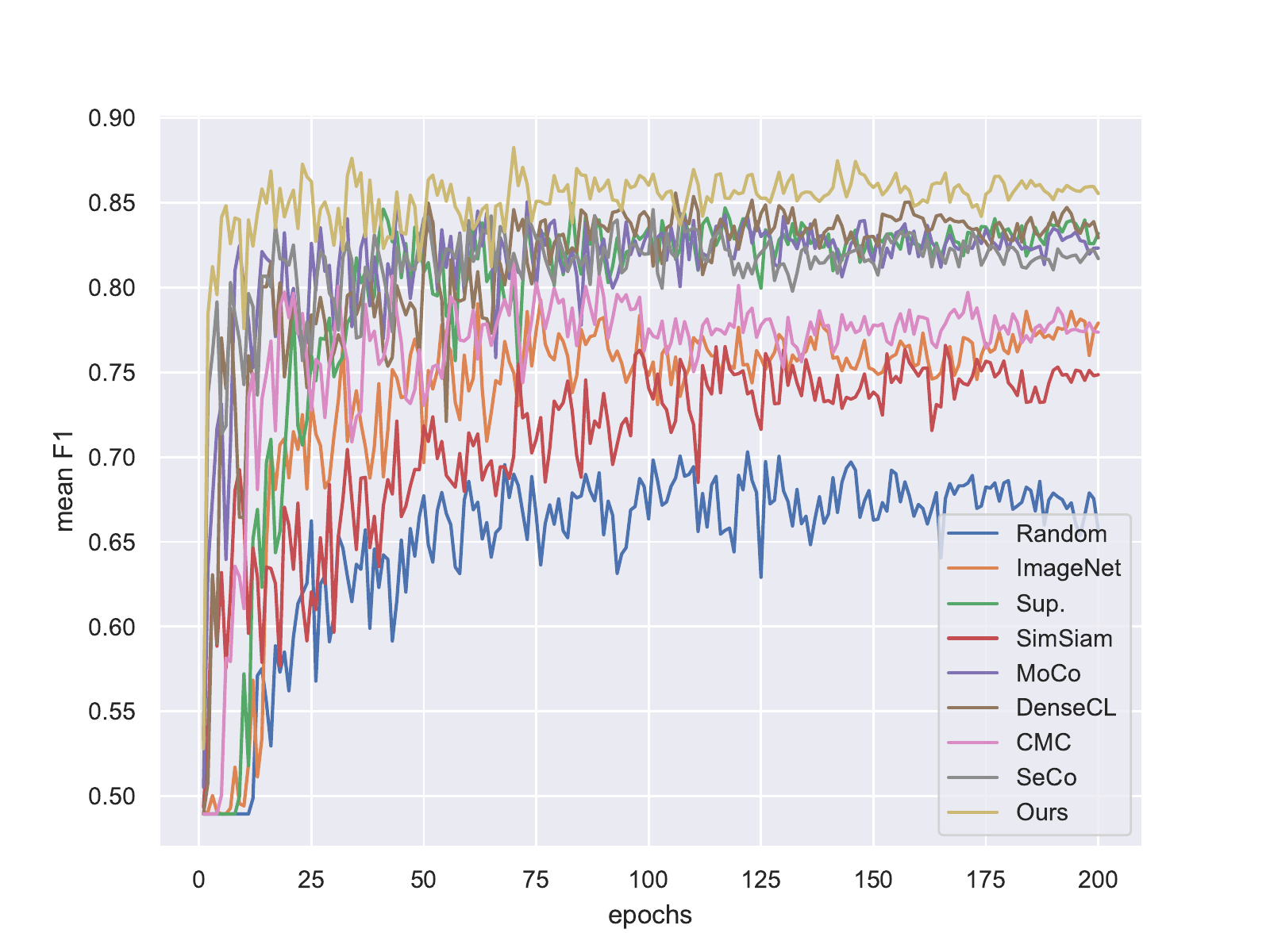}
  \centerline{(a) LEVIR-CD}
\end{minipage}
\begin{minipage}[t]{0.33\linewidth}
\centering
\includegraphics[width=\textwidth]{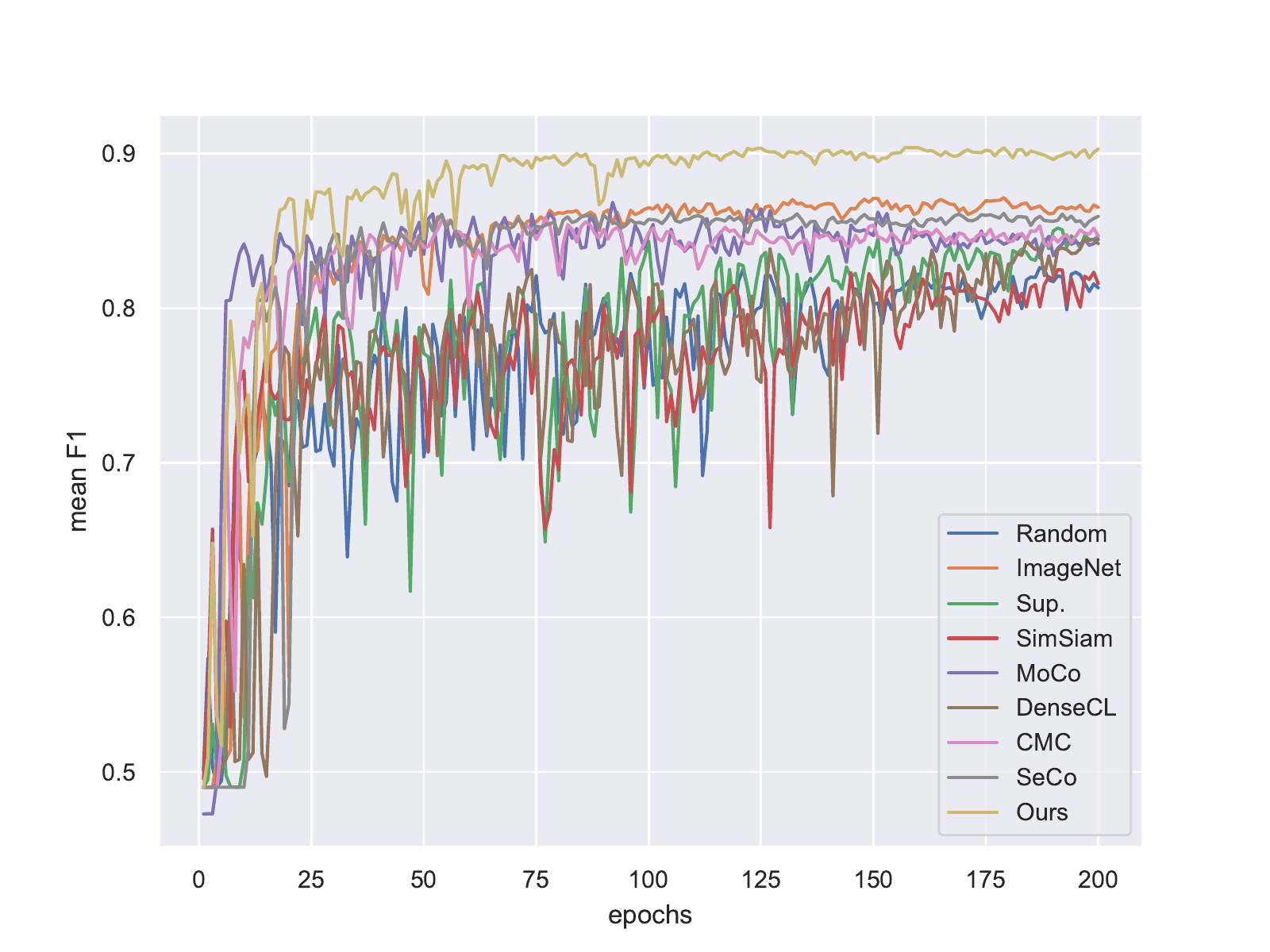}
  \centerline{(b) WHU-CD}
\end{minipage}
\begin{minipage}[t]{0.33\linewidth}
\centering
\includegraphics[width=\textwidth]{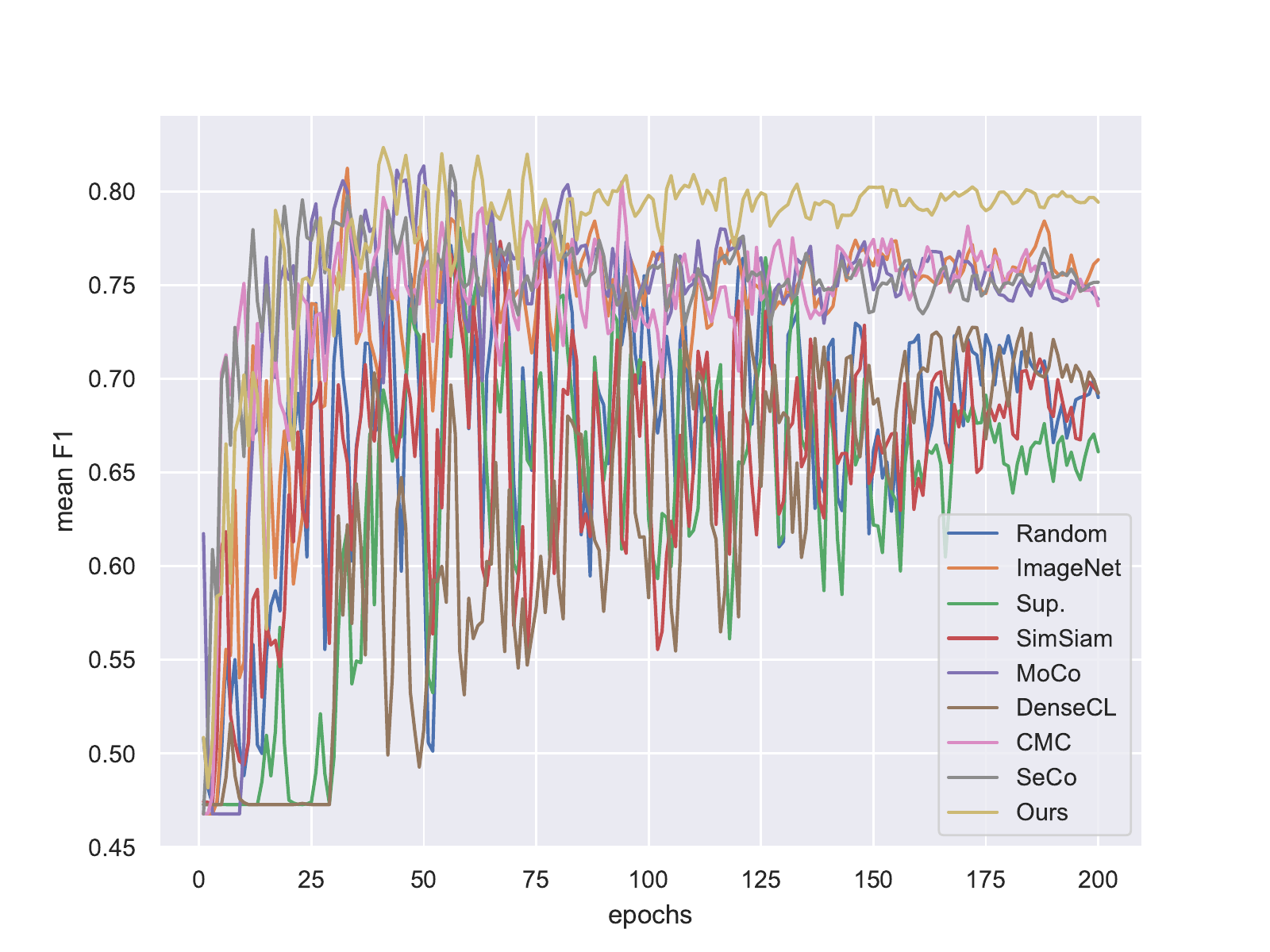}
  \centerline{(c) Guangzhou-CD}
\end{minipage}
\caption{Accuracy of CD models for each training epoch. The mean F1-score is reported.}
\label{fig:training_acc}
\end{figure*}

\textbf{Better convergence}.
We further give validation curves on the downstream CD task to show the transferability of each pre-trained model.
Fig. \ref{fig:training_acc} depicts the curve of validation accuracy (mean F1-score) during the training phase using 5\% LEVIR-CD/WHU-CD/Guangzhou-CD training data. Different colors denote different pre-training methods for a better view. It can be observed that our SaDL outperforms its counterparts in terms of stability and effectiveness. The comparison results indicate that our method provides better model initialization, which accelerates convergence and improves fine-tuning performance on the downstream CD task.

\textbf{Computational complexity}.
We also report the computational complexity of our method and other compared SSL methods. For a fair comparison, we test all the methods on a computing server equipped with an NVIDIA RTX 3090 GPU. Table \ref{tab:model_efficiency} lists the number of model parameters (Params.), floating-point operations per second (FLOPs), and GPU inference time of each method. We set the size of the image input to the model to $256 \times 256 \times 3$. The reported time is the average time consumption of each image over $100$ forward iterations given a random input sample of a batch size of $64$. The results show that our method requires relatively higher FLOPs (about $1.5$ times) than counterparts that have two views. It is because our SaDL needs to calculate three views each iteration. We can also observe that our method has fewer FLOPs than SeCo, but takes slightly more time. It may be due to the additional time consumption of our online view synthesis and masked sampling process. Nevertheless, our computational overhead is reasonable and acceptable.

\begin{table}
    \centering
    \caption{Computational complexity comparison. We report the number of model parameters (Params.), floating-point operations per second (FLOPs) and GPU inference time. The input image to the model has a size of $256\times 256\times 3$.}
    \begin{tabular}{c|ccc}
    \toprule
    Model & Params. (M) & FLOPs (G) & Time (ms) \\
    \midrule
        SimSiam \cite{Chen2020j} & 12.49 &  4.76 &  1.04 \\
        MoCo-v2 \cite{Chen2020d} & 11.24 &  4.76 &  1.92 \\
        DenseCL \cite{Wang2021a}  & 11.69 &  4.76  &  2.66 \\
        CMC \cite{Cha2021} & 22.48 & 4.66 &  1.55   \\
        SeCo \cite{Manas2021} & 12.16 & 9.52 &  3.62\\
    \midrule
        Ours & 15.09 & 7.83 & 4.17 \\
    \bottomrule
    \end{tabular}
    \label{tab:model_efficiency}
\end{table}

\subsection{Ablation Studies}
\label{ssec:ablation}

We conduct ablation studies on the three key components of our proposed SaDL: 1) dense representation learning via masked-based point sampling (MS), 2) learning discriminative features via semantic discrimination (SD), and 3) learning consistent foreground features via background swap (BS). We start from the SimSiam \cite{Chen2020j} with an FPN-based encoder, which learns consistent features between two artificially augmented views of the same image, as our baseline. We incrementally add the three key components to the baseline to evaluate their respective gains to the performance of the model. All the above models are trained on our pre-training dataset for 200 epochs and then transferred to the downstream CD task. Downstream experiments are performed under the 5\% data regime of the LEVIR-CD/WHU-CD/Guangzhou-CD datasets.

\textbf{Ablation on masked sampling (MS)}. Tab. \ref{tab:ablation-table} shows that our dense representation learning via sampling multiple points for each semantic category brings consistent and significant performance improvements across the three datasets, compared to the baseline. It indicates that our MS introduces much more valuable information to representation learning than the traditional global average pooling-based counterpart that loses spatial details. Moreover, we make a comparison to the masked pooling-based approach \cite{Chen2022c}, which pools features belonging to each semantic region into one semantic vector by leveraging the semantic mask. Simply constraining each semantic vector invariant across views brings small improvement, compared to the global pooling approach (baseline). We can observe that our sampling-based method is far superior to the masked pool one. The multiple pixel correspondences across views through masked sampling provide more diverse supervision of spatial accordance, thus benefiting representation learning.

\textbf{Ablation on semantic discrimination (SD)}. To make full use of the per-pixel semantic information, we employ an additional loss term, namely semantic dissimilarity loss, to learn discriminative features that can distinguish pixels from different semantic regions.
Tab. \ref{tab:ablation-table} shows that our SD induces considerable improvements in the F1-score of the CD model across the three datasets. 
The empirical results indicate that our semantic discrimination can significantly improve the model transferability. 
It may attribute to the role of the semantic mask in our SSL framework. We include the per-pixel semantic supervision into our pre-training in a contrastive manner. Different from directly mapping the image to the label, we learn an embedding space where pixels from the foreground (i.e., buildings) and the background are pushed apart. The resulting model obtains discriminative features that could distinguish land covers of interest from backgrounds, thus benefiting the recognition of change of interest in the downstream CD task.

\textbf{Ablation on background swap (BS)}. Apart from two views by artificial augmentations, we propose a third view via swapping the background of the current image from that of another one. This incurs an additional cross-view similarity loss, i.e., learning invariant foreground features between view 1 and view 3. Quantitative results in Tab. \ref{tab:ablation-table} demonstrate that our BS further brings consistent improvements in the performance of the pre-trained CD model on the three CD datasets. It indicates that the learned foreground representations invariant to background changes benefit the downstream change detection for the land covers of interest. Our background swap can be viewed as an approach to synthesizing unconcerned changes in a scene. Such design may enforce the model to focus more on changes of interest related to the foreground, but not irrelevant changes within the background. To further verify the effectiveness of the introduced third view, we also compare the natural augmentation, i.e., augmenting the current view with a registered image from another temporal. For a fair comparison, we apply the same loss for constraining consistent foreground representations between view 1 and view 3. Tab. \ref{tab:thirdview} demonstrates our synthetic view consistently outperforms the natural counterpart. While natural augmentation provides rich illumination changes, undesired real foreground changes could also occur, which may hinder representation learning.

\begin{table}
    \centering
    \caption{Ablation studies of our SaDL, including 1) dense representation learning via masked sampling (MS), 2) learning discriminative features via semantic discrimination (SD), and 3) learning consistent foreground features via background swap (BS). The F1-score on the LEVIR-CD/WHU-CD/GZ-CD test set is reported. The CD model is trained under the data regime of 5\%.}
    \resizebox{0.5\textwidth}{!}{
    \begin{tabular}{c|ccc|ccc}
   \toprule
    & MS & SD & BS & LEVIR & WHU & GZ  \\
    \midrule
    Baseline & $\times$ & $\times$ & $\times$ & 60.69 &  64.53 & 48.24  \\
    +masked pool & $\times$ & $\times$ & $\times$ & 63.09 & 66.03 & 48.64  \\
    \midrule
    Ours & $\checkmark$ & $\times$ & $\times$ & 70.21 & 71.90  & 54.15 \\
    Ours & $\checkmark$ & $\checkmark$ &$\times$ & 77.74 & 77.43 & 64.21   \\
    Ours & $\checkmark$ & $\checkmark$ &$\checkmark$ & \textbf{79.44} & \textbf{80.34} &  \textbf{65.89} \\
    \bottomrule
    \end{tabular}}
    \label{tab:ablation-table}
\end{table}

\begin{table}
    \centering
    \caption{Effect of the third view. The F1-score and IoU on the LEVIR-CD/WHU-CD/Guangzhou-CD test set are reported. The CD model is trained under the data regime of 5\%.}
    \resizebox{0.5\textwidth}{!}{
    \begin{tabular}{c|cc|cc|cc}
   \toprule
    \multicolumn{1}{c}{} &
    \multicolumn{2}{c|}{\textbf{LEVIR-CD}} &
    \multicolumn{2}{c|}{\textbf{WHU-CD}} &
    \multicolumn{2}{c}{\textbf{Guangzhou-CD}} 
    \\
     & F1 & IoU  & F1 & IoU & F1 & IoU   \\
    \midrule
    base & 77.74 & 63.58 &  77.43 & 63.17 & 64.21 & 47.29\\
    + natural & 78.32 & 64.37 &  78.86 & 65.10  &  62.99 & 45.98\\
    \midrule
    ours & \textbf{79.44} & \textbf{65.90} & \textbf{80.34} & \textbf{67.15} &  \textbf{65.89} & \textbf{49.13} \\
    \bottomrule
    \end{tabular}}
    \label{tab:thirdview}
\end{table}

\subsection{Parametric Analysis}
\label{subsec:parameter}

\textbf{Sampled point number}. The number of sampled points $N$ for each semantic category is an important hyperparameter. We test different $N\in \{1,4,16,64\}$ to analyze its effect on the performance of the downstream fine-tuning performance. Tab. \ref{tab:samplenumber} reports the F1-score/IoU of the pre-trained CD model trained on a 5\% data regime of the LEVIR-CD/WHU-CD/Guangzhou-CD datasets. We can observe an overall improvement in the F1 score of the CD model on the three datasets when increasing the sample number from 1 to 16. It is because that multiple sampled points from different spatial locations introduce more spatial-diverse information for both learning cross-view invariant and semantically discriminative features, thus facilitating the dense change recognition in the downstream CD task. We can also observe a drop in F1-score when further increasing $N$ from 16 to 64. It may be because the over-sampled points nearby the edge of the semantic regions could bring difficulty in training semantically discriminative features. Therefore, we set $N$ to 16.

\begin{table}
    \centering
    \caption{Effect of sampled point number $N$. The F1-score and IoU on the LEVIR-CD/WHU-CD/Guangzhou-CD test set are reported. The CD model is trained under the data regime of 5\%.}
    \resizebox{0.5\textwidth}{!}{
    \begin{tabular}{c|cc|cc|cc}
   \toprule
    \multicolumn{1}{c}{} &
    \multicolumn{2}{c|}{\textbf{LEVIR-CD}} &
    \multicolumn{2}{c|}{\textbf{WHU-CD}} &
    \multicolumn{2}{c}{\textbf{Guangzhou-CD}} 
    \\
    N & F1 & IoU  & F1 & IoU & F1 & IoU   \\
    \midrule
    1 & 74.00 & 58.73 &  78.49 & 64.59 & 63.51 & 46.53\\
    4 & 75.98 & 61.26 &  80.23 & 66.98  & \textbf{66.07} & \textbf{49.33}\\
    16 & \textbf{79.44} & \textbf{65.90} & \textbf{80.34} & \textbf{67.15} &  65.89 & 49.13 \\
    64 & 76.12 & 61.45 &  76.34 & 61.73  &  64.84 & 47.97  \\
    \bottomrule
    \end{tabular}}
    \label{tab:samplenumber}
\end{table}

\begin{table}
    \centering
    \caption{Effect of the number of pre-training epochs. The F1-score and IoU on the LEVIR-CD/WHU-CD/Guangzhou-CD test sets are reported. The CD model is trained under the data regime of 5\%.}
    \resizebox{0.5\textwidth}{!}{
    \begin{tabular}{c|cc|cc|cc}
   \toprule
    \multicolumn{1}{c}{} &
    \multicolumn{2}{c|}{\textbf{LEVIR-CD}} &
    \multicolumn{2}{c|}{\textbf{WHU-CD}} &
    \multicolumn{2}{c}{\textbf{Guangzhou-CD}} 
    \\
    epochs & F1 & IoU  & F1 & IoU & F1 & IoU   \\
    \midrule
    0 & 65.18 & 48.35 & 73.52 & 58.12 & 57.86 & 40.71 \\
    \midrule
    5 & 73.68 & 58.33 &  75.39 & 60.51 & 60.54 & 43.41\\
    20 & 76.48 & 61.92 & 79.17 & 65.52 & 65.61 & 48.82\\
    100 &  77.14 & 62.79 & 79.76 &  66.34  &  64.57 & 47.68\\
    200 & 79.44 & 65.90 & \textbf{80.34} & \textbf{67.15} &  \textbf{65.89} & \textbf{49.13} \\
    400 & \textbf{79.59} & \textbf{66.10} & 80.06 & 66.75 &  64.37 & 47.47  \\
    \bottomrule
    \end{tabular}}
    \label{tab:epochs}
\end{table}

\textbf{Pre-training epochs number}. The epochs number $N_{e}$ of pre-training affects its fine-tuning performance on the downstream task. We train our SaDL using varying epochs numbers $N_{e} \in \{5, 20, 100, 200, 400\}$. Tab. \ref{tab:epochs} reports the F1-score/IoU of the pre-trained CD model on the LEVIR-CD/WHU-CD/Guangzhou-CD datasets under a 5\% data regime.
Note that $N_{e}=0$ denotes the ImageNet pre-training. We can observe a significant improvement in the F1-score even with a very small number of pre-training epochs (e.g., 5), compared to the baseline ($N_{e}=0$). The performance of the pre-trained CD model further improves with longer training (e.g., from 5 to 200 epochs). We can also obverse a negligible improvement or even some drops in the F1-score of the CD model on the downstream tasks when further increasing the $N_{e}$ from 200 to 400. A pre-training with too many training epochs may overfit the pre-training data and incur poorer model transferability. Therefore, we set $N_{e}$ to 200.

\begin{figure}
        \centering
        \includegraphics[width=0.5\textwidth]{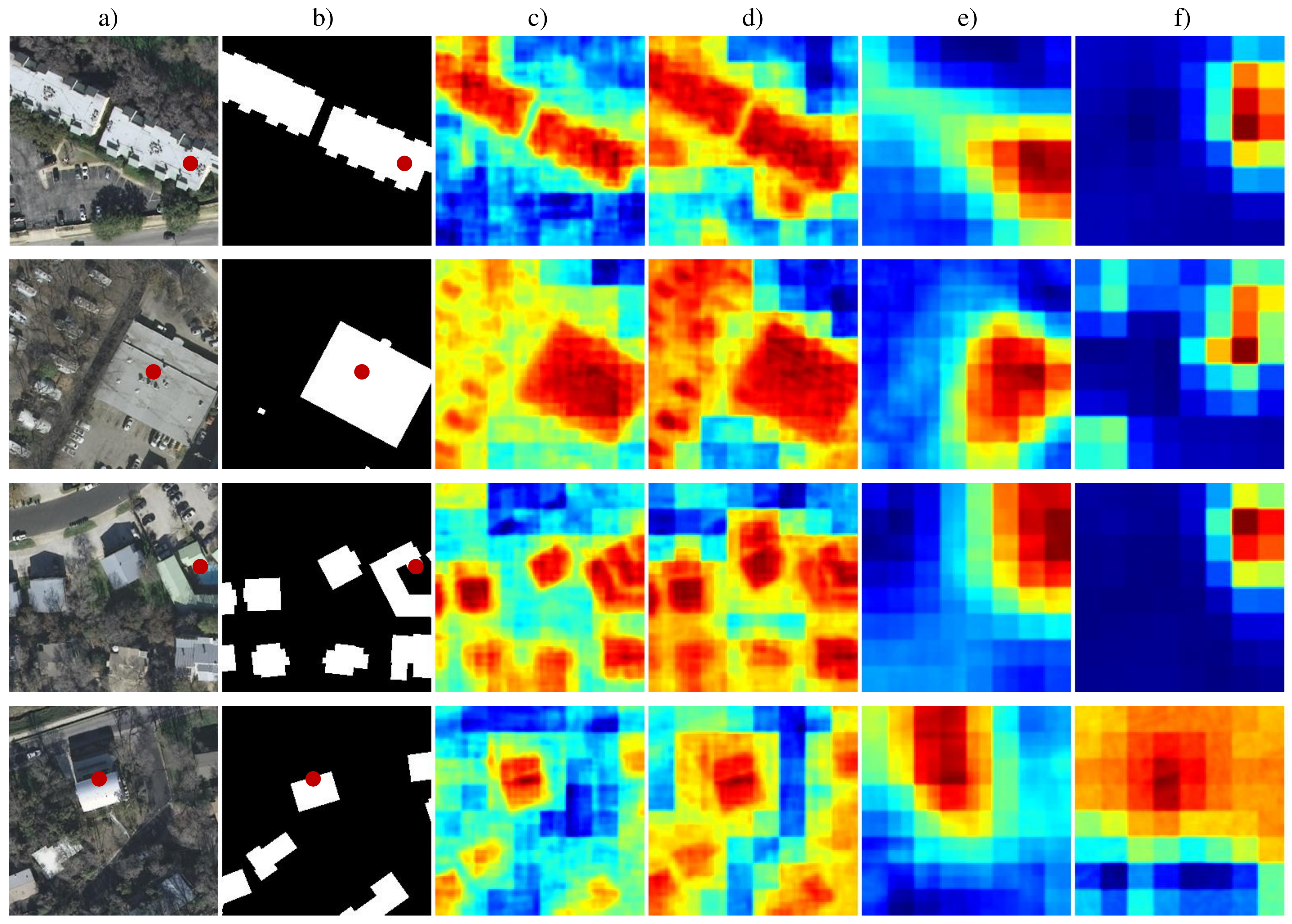}   
        \caption{Visualization of self-similarity map of our SaDL on the pre-training dataset. Given one selected point (viewed in red) in the a) image, we give visualization results of the ablation versions of our model, including f) baseline, e) ours with masked sampling (MS), d) ours with MS and semantic discrimination (SD), c) ours with MS, SD, and background swap (BS). The semantic mask b) is shown for a better view.}
        \label{fig:selfsim}
\end{figure}

\begin{figure*}
        \centering
        \includegraphics[width=\textwidth]{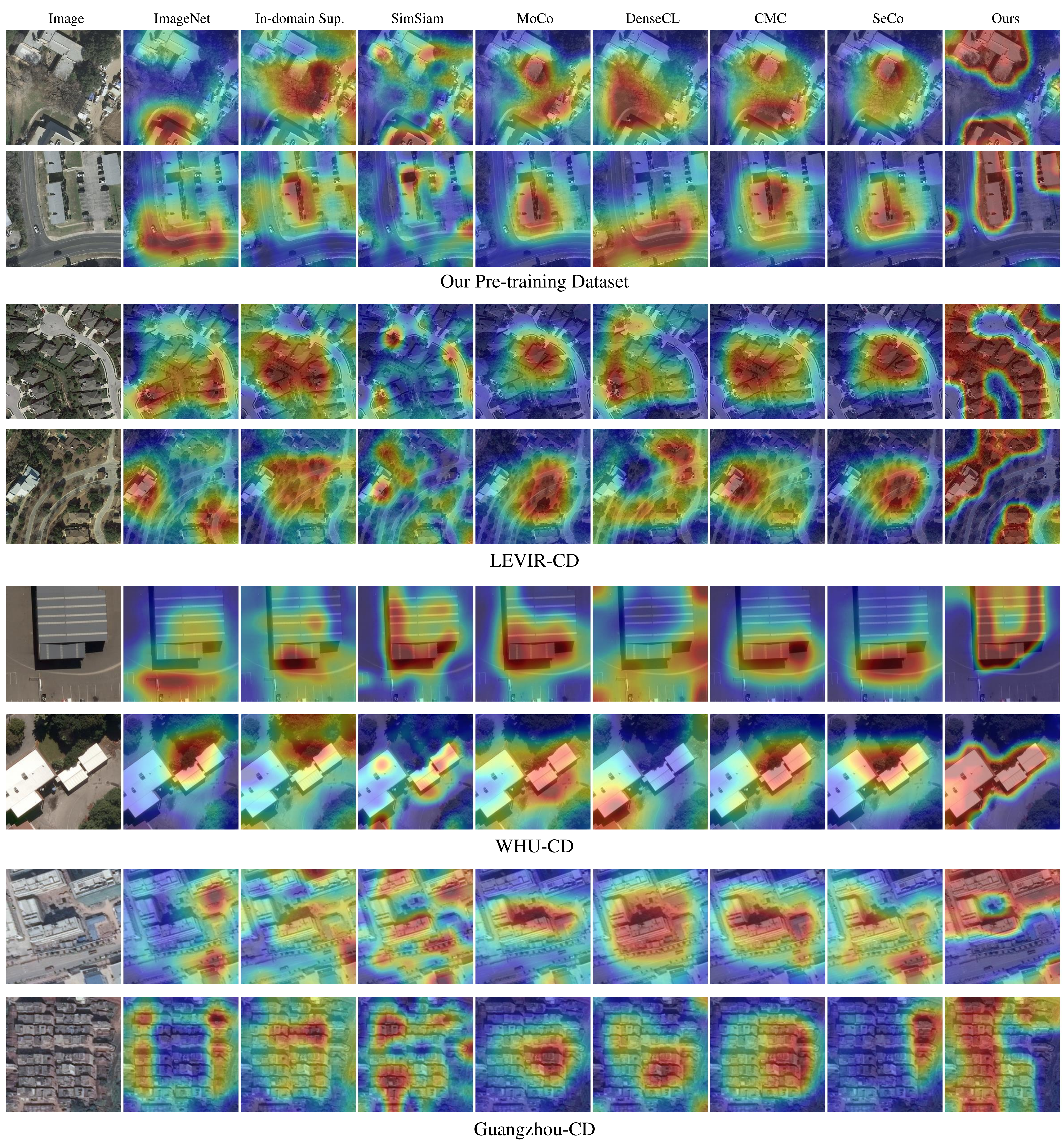}   
        \caption{Class activation maps of different pre-trained models. We test images from our pre-training dataset and the downstream LEVIR-CD/WHU-CD/Guangzhou-CD datasets.}
        \label{fig:cam}
\end{figure*}

\subsection{Feature Visualization}

\subsubsection{Encoder visualization}
We incorporate semantic supervision into a self-supervised framework to learn semantically meaningful per-pixel representations. To better inspect the semantic relation of the learned representations, we visualize the self-similarity map of the output dense features from our encoder. 

\textbf{Feature self-similarity map}. For one selected point on an image, we can obtain one self-similarity map by calculating feature correlations between the selected point and all the pixels in the image. Formally, given the per-pixel features $\mathbb{X} \in \mathbb{R}^{HW \times C}$ and one selected point feature $\mathbf{x}=\mathbf{X}[p] \in \mathbb{R}^{C}, p=(u,v)$, the self-similarity map $SM \in \mathbb{R}^{HW}$ of the point $p$ is given by their dot product followed by softmax on the channel of $HW$, i.e., $SM=\text{softmax}(X\cdot x)$.

Fig \ref{fig:selfsim} shows some examples of self-similarity maps of the selected foreground points (viewed in red) on images in our pre-training dataset. Red denotes higher similarity values and blue denotes lower values. From Fig. \ref{fig:selfsim} c), we can observe that the selected foreground point shows strong correlations with pixels of the same semantic category (building). It indicates that our SaDL can obtain high-resolution high-level features that reveal semantic relations between pixels.

Fig. \ref{fig:selfsim} also gives visualization results for other ablation versions of the proposed model, including our baseline [f)], baseline + MS [e)], and baseline + MS + SD [d)]. We can observe that ours with MS can highlight pixels nearby the selected point and shows a relatively more accurate location response than the baseline.
It implies the effectiveness of the proposed dense representation learning based on masked sampling over the traditional average pooling approach. 
We can also observe our ablation [d)] highlights more accurate foreground regions than that of e). It indicates our introduced SD incurs discriminative representations that well distinguish the foreground and the background. Moreover, our full SaDL model [c)] obtain more refined visualization results by further adding BS to learn consistent foreground features invariant to background changes.

\subsubsection{Backbone visualization}
For a fair comparison, we only transfer backbone (i.e., ResNet-18) parameters of our SaDL on the downstream CD model, like other pre-trained models. To interpret what concepts are learned by the networks, we show the backbone visualization of each pre-trained model by utilizing network visualization techniques - Class Activation Maps (CAM) \cite{Wang2020f}. Basically, a CAM is the weighted sum of each channel activation map from a certain layer in the network. Here, we visualize the last convolution layer of the backbone. 

Fig. \ref{fig:cam} shows the comparison of CAM for different pre-trained models. Red denotes higher attention values and blue denotes lower values. We test images from our pre-training dataset as well as the three downstream CD datasets. We can observe that the CAM of our model shows high concentrations at the foreground objects in the image with a satisfactory localization accuracy. It is notable that our model can not only attend to land covers of interest on the pre-training images but also perform well on the downstream images with considerable domain gaps (e.g., differences in spatial resolution and object style) from the pre-training data. For instance, the high-density buildings (see the last row in Fig. \ref{fig:cam}) in the Guangzhou-CD dataset are hardly present in the pre-training dataset. It indicates that our proposed pre-training could extract transferable semantically meaningful representations with well generalization ability.

\section{Discussion}
\label{sec:discussion}
We present semantic-aware representation learning for change detection in high-resolution remote sensing images. One major challenge of change detection is how to improve the model generalization ability under the insufficient training labeled data regime. The issue of label insufficiency comes from two aspects, 1) imaging and land-cover diversity making it impossible to collect training samples covering real-world conditions, and 2) extensive labor to collect bitemporal images containing changes and to annotate per-pixel labels. We leverage existing freely available data with semantic information in the remote sensing community to learn a pre-trained model to benefit the downstream CD task. Considering the key of CD is to recognize the real changes and exclude other irrelevant changes, we incorporate such CD task priors in the pre-training phase by learning semantically meaningful representations that could distinguish foreground land-covers from other backgrounds, and that are invariant to irrelevant changes, e.g., illumination differences and unconcerned land-cover changes. We explore semantic-supervised pre-training in a contrastive manner, which is different from traditional supervised pre-training that learns the mapping from the image to the label or other self-supervised learning methods that lack of semantic supervision. Compared to several existing pre-trained methods, our method significantly improves the CD data efficiency (see Fig. \ref{fig:predictions}) and could extract semantically meaningful representations with good generalization ability even on the unseen domain (see Fig. \ref{fig:cam}). It indicates our pre-trained method can learn transferable discriminative features that benefit change recognition and incur better model generalization in the downstream CD task.

\section{Conclusion}
\label{sec:conclusion}

In this paper, we proposed a semantic-aware pre-training method for remote sensing image change detection. We incorporate the semantic information into a self-supervised learning framework to enhance the feature discrimination ability. Instead of manipulating the image-level features, we constrain pixel-level representation consistency based on sampling multiple spatially aligned points across views by introducing location correspondences. We employ an additional loss term to learn discriminative features that can distinguish foreground land covers and other backgrounds. Apart from two views via artificial augmentations, we generate a third view via background swap to learn consistent presentations invariant to irrelevant changes. Extensive experiments on three downstream CD datasets verify the effectiveness of the proposed method. Our SaDL significantly outperforms ImageNet pre-training, in-domain supervised pre-training, and several SSL pre-training methods, especially in small data regimes. The empirical results indicate that SaDL can well alleviate the labeled data insufficiency in CD. Notably, we can achieve comparable or even better results using only 20\% training data than the random initiation baseline using 100\% data. The feature visualization results also demonstrate that our method can extract semantically meaningful representations with well generalization ability on different downstream images with domain discrepancies from the pre-training data.

Modern machine learning-based models typically suffer from spectral variabilities \cite{Hong2019a} in remote sensing data, caused by different imaging conditions, including illuminations, atmospheric effects, and instrumental configurations. Our proposed pre-trained model could give consistent representations invariant to irrelevant changes (e.g., illumination and unconcerned land cover changes), to benefit the downstream CD task. In the future, we will explore effective ways to explicitly model more kinds of imaging conditions in our pre-training framework to better handle the spectral variabilities.
Other future directions include 1) sampling strategies. Instead of treating each pixel equally, we could leverage the guidance map generated by edge clues or network attention maps to sample important pixels that may better benefit the representation learning. 2) exploring various pre-training backbones, e.g., more kinds of CNN structures and even other vision transformer architectures. 3) extending our semantic-aware pre-training to distinguish more land cover types, which may benefit the multi-class semantic change detection task.





\ifCLASSOPTIONcaptionsoff
  \newpage
\fi

{\small
\bibliographystyle{IEEEtran}
\bibliography{references}
}

\end{document}